 \documentclass[final,5p,times,twocolumn]{elsarticle}

\usepackage{amssymb}
\usepackage{amsfonts}
\usepackage{amsthm}
\usepackage{graphics} 
\usepackage{epsfig} 
\usepackage{amsmath} 
\usepackage{subfigure}
\usepackage{multirow}
\usepackage{makecell}
\usepackage{multirow}
\usepackage{graphicx}
\usepackage{lscape}
\usepackage{multicol} 
\usepackage{graphicx, epstopdf}
\usepackage[table,xcdraw]{xcolor}
\usepackage{xspace,epsfig,url}
\usepackage{enumerate}
\usepackage{float}
\usepackage{epsf}
\usepackage{psfrag}
\usepackage{verbatim}
\usepackage[all]{xy}
\usepackage{color,soul}
\usepackage{bm}
\usepackage{comment}
\usepackage{cases}
\usepackage{hyperref}
\usepackage{physics}
\usepackage{amssymb}
\usepackage{pifont}
\usepackage{array} 
\usepackage[ruled,vlined,linesnumbered]{algorithm2e}

\newcommand\todo[1]{\textcolor{red}{#1}}

\newcommand{\expnumber}[2]{{#1}\mathrm{E}{#2}}

\journal{ArXiv Preprint}

\begin{document}

\begin{frontmatter}

\title{Enhanced Optimization Strategies to Design an Underactuated Hand Exoskeleton}

\author[inst1]{Baris Akbas}
\author[inst2,form]{Huseyin Taner Yuksel}
\author[inst2,form]{Aleyna Soylemez}
\author[inst2]{Mine Sarac}
\author[inst2,corr]{Fabio Stroppa}

\affiliation[inst1]{organization={Istanbul Technical University},
            addressline={Reşitpaşa, İTÜ Ayazağa Kampüsü Rektörlük Binası}, 
            city={Istanbul},
            postcode={34467}, 
            country={Turkey}}
\affiliation[inst2]{organization={Kadir Has University},
            addressline={Cibali, Kadir Has Cd., Fatih}, 
            city={Istanbul},
            postcode={34083}, 
            country={Turkey}}
\affiliation[form]{Formerly}
\affiliation[corr]{Corresponding Author: fabio.stroppa@khas.edu.tr}

\begin{abstract}
Exoskeletons can boost human strength and provide assistance to individuals with physical disabilities. However, ensuring safety and optimal performance in their design poses substantial challenges. This study presents the design process for an underactuated hand exoskeleton, first including a single objective (maximizing force transmission), then expanding into multi-objective (also minimizing torque variance and actuator displacement). The optimization relies on a  Genetic Algorithm, the Big Bang-Big Crunch Algorithm, and their versions for multi-objective optimization.
Analyses revealed that using Big Bang-Big Crunch provides high and more consistent results in terms of optimality with lower convergence time. In addition, adding more objectives offers a variety of trade-off solutions to the designers, who might later set priorities for the objectives without repeating the process -- at the cost of complicating the optimization algorithm and computational burden. These findings underline the importance of performing proper optimization while designing exoskeletons, as well as providing a significant improvement to this specific robotic design.
\end{abstract}

\begin{keyword}
Hand Exoskeleton \sep Rehabilitation Robotics \sep Evolutionary Computation \sep Multi-Objective Optimization \sep Design Optimization
\end{keyword}

\end{frontmatter}



\section{Introduction}
Exoskeletons are wearable robotic devices that can improve users' functional abilities during physically challenging tasks~\cite{poliero2020applicability,butzer2021fully} or assist/rehabilitate them in case of functional disabilities~\cite{stroppa2017robot,huang2015design,pons2010rehabilitation}. They are usually designed to be worn on a specific body location; thus, their proper design is crucial for force transmission efficiency but also, and most importantly, for users' safety~\cite{sarac2019design}. For an exoskeleton to be safe, the anatomic and mechanical joints must be aligned during possible (intentional or unintentional) movements of users~\cite{van2015considerations,fisahn2016effectiveness}. In addition, the exoskeleton design must be optimized to provide the highest level of well-distributed output forces through small and lightweight actuators. 


The engineering design is usually performed through mathematical optimization, searching for the best element (\textit{optimum}) within a set of alternatives (decision variables' values) based on one or more specific criteria (\textit{objective}s)~\cite{sioshansi2017optimization, statnikov2012multicriteria, andersson2000survey}.
Traditional optimization methods rely on numerical and calculus-based techniques~\cite{bonnans2006numerical} and might be challenging for exoskeleton design~\cite{stroppa2023optimizing} due to properties such as non-discrete domains, multimodality, discontinuity~\cite{norde2000characterizing}, and the need for reliability~\cite{dizangian2015reliability} and robustness~\cite{nomaguchi2016robust}.
Alternatively, the nature-inspired Evolutionary Computation (EC) methods -- particularly Evolutionary Algorithms (EAs) -- are well-known and effective in dealing with engineering optimization problems~\cite{dumitrescu2000evolutionary, wang2017autonomous} and often with exoskeleton design~\cite{stroppa2023optimizing, datta2015analysis} and other robotic devices~\cite{stroppa2024optimizing}. 

\begin{figure}[t!]
    \centering
    \includegraphics[width=0.8\linewidth]{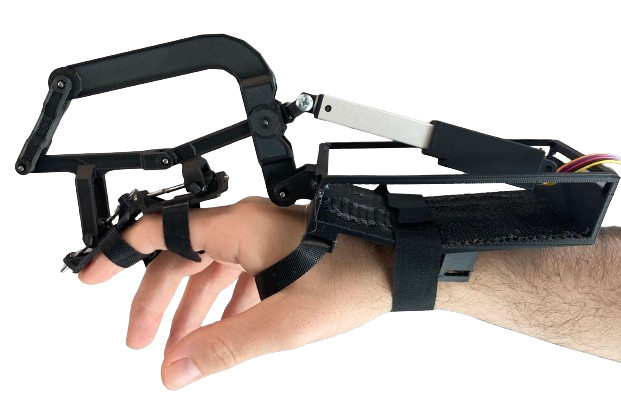}
    \caption{The optimized device for the index finger component of U-HEx. 
    }
    \label{fig:uhex_new}
\end{figure}

Previously, we proposed optimizing an underactuated hand exoskeleton (U-HEx~\cite{sarac2017design}, shown in Fig.~\ref{fig:uhex_new}) using EAs, in which we retrieved the best set of link lengths (decision variables) that maximize the output force transmission (objective function)~\cite{akbas2024impact}. 
We specifically chose EAs over other nonlinear numerical methods due to their robustness in handling black-box optimization and their ability to efficiently explore the design space of our underactuated hand exoskeleton without requiring explicit gradients or analytical formulations for the objective functions -- especially considering that the objective functions come from a motion simulation of the kinematic chain.
While the first design of U-HEx was optimized with a naive-deterministic search approach, which iteratively analyzed each possible combination of design parameters (i.e., ``brute force''), we employed a Genetic Algorithm (GA)~\cite{goldberg1987genetic,goldberg1990real} and the Big Bang-Big Crunch algorithm (BBBC)~\cite{erol2006new}. Comparative analyses revealed that both EAs yielded more consistent and optimal solutions than brute force in a significantly shorter time. This allowed us to extend the search space by including more decision variables (i.e., three more link lengths), significantly improving U-HEx's efficiency.

Yet, our previous work had a few limitations. Firstly, we only focused on a 
single objective function. Secondly, 
we did not constrain the desired actuator movements to reach a natural range of motion for the user's hand. In other words, we have found an \textit{optimal} set of link lengths with no prior assumption on the desired actuator displacements. This assumption cannot be overlooked, since our design involves a linear actuator with a physical limit to how much it could move from its base 
-- so that it can fit on top of the hand~\cite{sarac2017design}. Therefore, it is crucial to minimize the desired range of actuator displacement to assist natural ranges of motion for the finger joints \textit{without} sacrificing the force transmission performance. On the other hand, we have previously constrained the ratio between the actuated finger joints in our optimization problem, while it would be preferred to be as close to 1 for well-distributed and safe force transmission. 


In this study, we have built up on our Single-Objective Optimization Problem (SOOP)~\cite{akbas2024impact} by improving U-HEx's design in two alternative design processes:

\begin{enumerate}
    \item we redefined the SOOP with an additional constraint to the actuator displacement based on the original design~\cite{sarac2017design}, and we hypothesize that this would result in a lower overall transmission and potentially different trends between optimization methods -- even though we have very effective optimization methods in hand ($H_1$); and
    \item we extended the SOOP to a Multi-Objective Optimization Problem (MOOP) by \textit{also} optimizing the balance between the torques applied to each finger joint and minimizing the required actuator displacement (i.e., turning constraints into objectives), and we hypothesize that treating constraints as further objectives would better accommodate the problem and lead to a trade-off set of different designs balancing the device's properties ($H_2$). 
\end{enumerate}



Our contributions involve \textit{(i)} providing a set of optimized designs for a state-of-the-art exoskeleton that was originally optimized with inefficient methods~\cite{sarac2017design}, \textit{(ii)} providing a walkthrough for a robotic design process that can be used as a guideline by future and less experienced robotic designers, \textit{(iii)} comparing different algorithms and strategies for MOOPs to determine which one is most suitable for our design problem, and \textit{(iv)} introducing two novel EC methods (NS-BBBC and SP-BBBC) suitable for robotic optimization, which are based on a state-of-the-art algorithm for single-objective optimization (i.e., BBBC). Table~\ref{tab:novelties} summarizes the different contributions on each work of the U-HEx device, comparing the current work with its state-of-the-art.

\begin{table}[h!]
\caption{Comparison of different works novelties on U-HEx}
\label{tab:novelties}
\centering
\renewcommand{\arraystretch}{1.2} 
\setlength{\tabcolsep}{3pt} 
\small
\begin{tabular}{|p{0.28\linewidth}|p{0.32\linewidth}|p{0.35\linewidth}|} 
\hline
\textbf{Reference} & \textbf{Novelty} & \textbf{Optimization Method} \\ \hline
Sarac et al. 2017~\cite{sarac2017design} & Kinematics and mechanical design & Brute Force \\ \hline
Akbas et al. 2024~\cite{akbas2024impact} & Optimized device based on maximum torque magnitude & Brute Force, GA, BBBC \\ \hline
Current Work & Comprehensive and broader optimization of the device on multiple criteria
& GA, BBBC, NSGA-II, SPEA2, NS-BBBC, SP-BBBC \\ \hline
\end{tabular}
\end{table}

The rest of the work is organized as follows: 
Sec.~\ref{sec:methods} defines the optimization design and provides a background on the methods we used; 
Sec.~\ref{sec:design_process_soop} describes the first optimization problem with a single objective (i.e., maximize the force transmission); 
Sec.~\ref{sec:design_process_moop} describes the second optimization problem with two more objectives (i.e., minimize the distribution of forces and actuator displacement);  
Sec.~\ref{sec:comparison} discusses the differences in the retrieved designs and the impact of each optimization process; and
Sec.~\ref{sec:conclusion} summarizes the work and presents future ideas.

\section{Exoskeleton Design Optimization} \label{sec:methods}

\subsection{U-HEx Kinematics} \label{sec:kinematics}

U-HEx in Fig.~\ref{fig:uhex_new} is a hand exoskeleton designed for rehabilitation~\cite{sarac2017design}. It was designed as a linkage structure with mechanical joints intentionally misaligned from the anatomical joints of the user. This design choice has multiple advantages. Firstly, the misalignment offers a much safer operation, since the mechanical device does not impose hard forces on the joints directly. Any sort of undesired or extreme force would be transmitted to the fingers in a more natural way without the risk of harming the user. Secondly, thanks to this misalignment, U-HEx can adapt its operation automatically to users with different finger sizes as well. Finally, since U-HEx joints do not need to be aligned with the anatomical joints strictly, it is noticeably easy to wear. 

The misalignments of U-HEx were designed carefully to achieve underactuation. A linear DC motor allows the device to close/open the finger in a direct manner. 
The idea behind underactuation is simple: the actuation is turned into motion at the first output joint until the first mechanical link gets in contact with an object. After the first link reaches an object and starts feeling the interaction forces, the spring-like joint coupling between output joints allows the mechanism to transmit the same actuation for the second output joint. Actuator forces are transmitted to output joints until the grasping is completed~\cite{Denizon2024, Yan2022}. From the point of hand exoskeletons, users' fingers as a part of the mechanism: anatomical coupling between finger joints defines the required spring-like behavior for the underactuation~\cite{Li2023, Gu2024, sarac2017design}. Then, efficient force transmission requires the use of additional passive joints, which can assist motion for finger joints without aligning mechanical joints to the finger joints. 

\begin{figure}[t!]
    \centering
\includegraphics[width=\columnwidth]{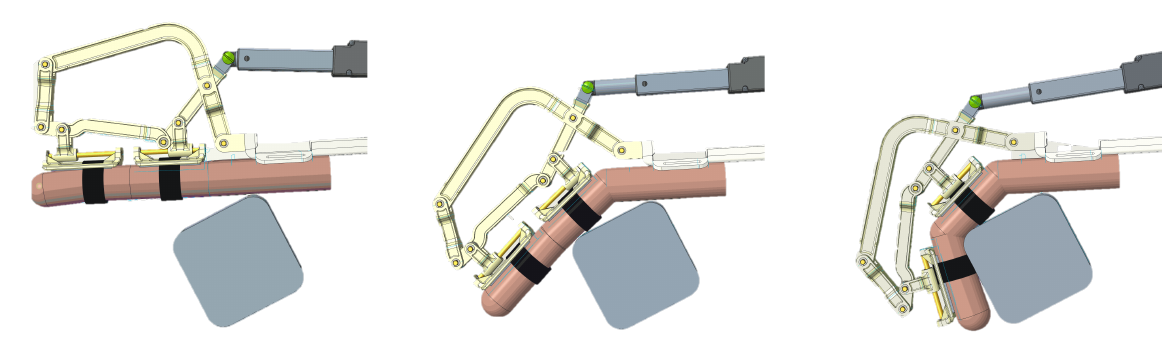}
    \caption{Underactuation concept of U-HEx during a grasping task: (left) initial pose with all finger joints extended, (mid) actuator force moves the MCP joint until the first finger phalange gets in touch with the object, (right) actuator force is transmitted to move the PIP joint when the first finger phalange touches the object and finally grasping task is completed when both phalanges touch the object}
    \label{fig:underact}
\end{figure}

Fig.~\ref{fig:underact} represents the operation flow of U-HEx. The finger component is connected to the user's finger and assists the finger in flexing until both finger joints are fully extended and the actuator stroke is at its minimum limit. In the beginning, as the actuator starts to move, the exoskeleton transmits forces to the proximal phalange and rotates the MCP joint. Encountering external forces along the proximal phalange caused by the physical interaction finalizes the rotation along the MCP joint. These forces transmit actuation forces to the intermediate phalange through the linkage-based mechanism with passive joints and rotating the PIP joint. Having two finger joints to control, the grasping task is completed when the PIP joint is rotated enough for the intermediate phalange to reach the object. Similarly, the extension of the finger starts by rotating the PIP joint first and the MCP joint later until the finger is totally extended and finger joints reach their physical limits. As a safety measure, the displacement stroke is designed to reach its minimum limit when both finger joints are extended to their natural limits.

\begin{figure}[t!]
    \centering
    \includegraphics[width=\columnwidth]{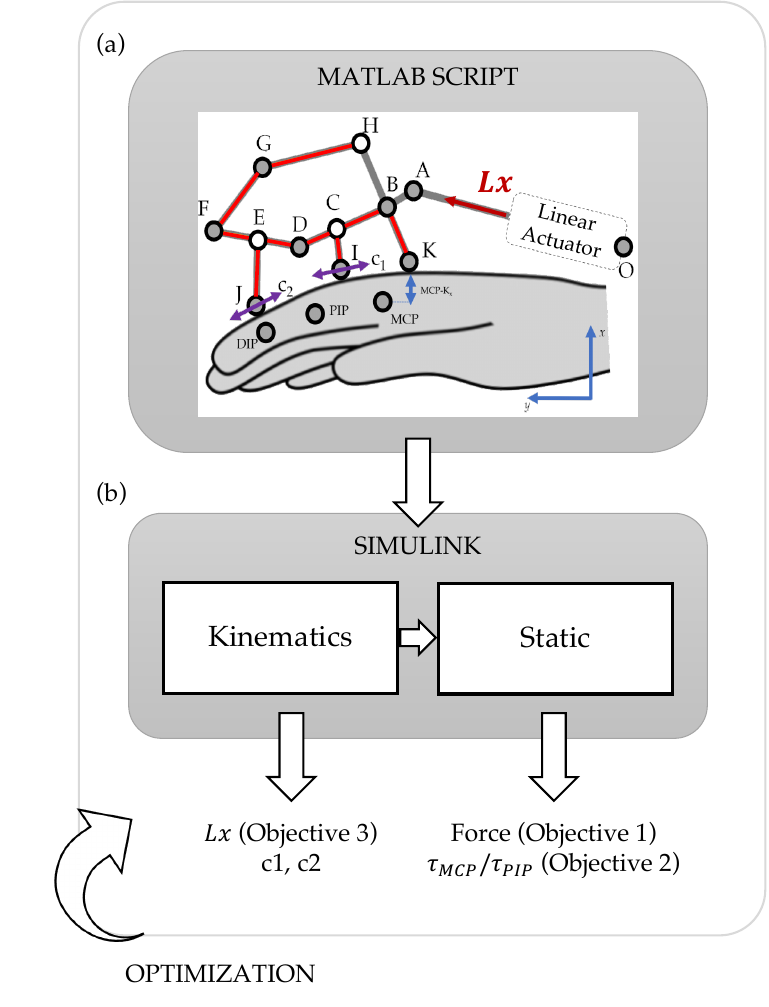}
    \caption{Optimization process for U-HEx: 
    for a given set of link lengths, a Simulink model is executed to compute numerical inverse kinematics and analytical statics as the finger joints are iterated from fully open to fully closed; while the optimization algorithms are implemented in MATLAB.}
    \label{fig:schema}
\end{figure}

Fig.~\ref{fig:schema}(a) represents the kinematic structure of U-HEx. Each gray dot represents a passive rotational joint (or anatomical joint) while empty circles ($H$, $E$, and $C$) are fixed points along rigid links with $90^o$ fixed angle. The system has only one linear actuator between the points $O$ and $A$. The exoskeleton is fixed on the hand from points $K$ and $O$ with variable lengths in the $x$ and $y$ directions. Finally, the exoskeleton is attached to the finger phalanges from points $I$ and $J$ with passive linear joint sliders represented as $c_1$ and $c_2$, respectively. 


U-HEx design is optimized to retrieve the optimal link lengths following the schema and design process summarized in Fig.~\ref{fig:schema}. The decision variables of the problem are the link lengths (depicted as red lines in the figure). The original U-HEx~\cite{sarac2017design} was optimized only for six link lengths: $\overline{BC}$, $\overline{CD}$, $\overline{DE}$, $\overline{EF}$, $\overline{FG}$, and $\overline{GH}$. due to the heavy computational requirements of brute force, which made it impossible to run the optimization with more link lengths. 
In our previous work~\cite{akbas2024impact}, 
we included an additional three link lengths ($\overline{BK}$, $\overline{CI}$, and $\overline{EJ}$) in the optimization problem, potentially creating a more optimal solution. 
The bounds of the decision variables (i.e., the range of the link lengths) are reported in Table~\ref{tab:bounds} and were chosen empirically~\cite{sarac2017design}. 

\begin{table}[h!]
\centering
\caption{Link Length Ranges / Decision Variable Bounds} 
    \label{tab:bounds}
    \resizebox{\columnwidth}{!}{%
    \begin{tabular}{cccccc}
    \multicolumn{6}{c}{Main Six Link Lengths (mm)}  \\$\overline{BC}$ & $\overline{CD}$ & $\overline{DE}$ & $\overline{EF}$ & $\overline{FG}$	& $\overline{GH}$ \\ \hline
        $38 \rightarrow 60$ & $10 \rightarrow 30$ & $15 \rightarrow 51$ & $15 \rightarrow 51$ & $27 \rightarrow 56$ & $64 \rightarrow 100$ \\ \hline
                   &           &            &           &            &           \\
    \multicolumn{6}{c}{Additional Three Link Lengths (mm)} \\
    \multicolumn{2}{c}{$\overline{BK}$} & \multicolumn{2}{c}{$\overline{CI}$} & \multicolumn{2}{c}{$\overline{EJ}$ } \\ \hline
    \multicolumn{2}{c}{$20 \rightarrow 50$} & \multicolumn{2}{c}{$10 \rightarrow 17$} & \multicolumn{2}{c}{$20 \rightarrow 50$} \\ \hline
    \end{tabular}
    }
\end{table}

Fig.~\ref{fig:schema} shows that at each step of the process, a new set of link lengths is assigned by the optimization method in MATLAB. This set is used to compute the numerical inverse kinematics~\cite{sarac2017design} as the finger joints are iterated from fully open ($0^o$) to fully closed ($80^o$ MCP; $90^o$ PIP) with a fixed-step solver. This process results in the corresponding movements around the joints (e.g., linear sliders $c_1$, $c_2$) and the desired actuator displacement ($L_{x}$). These movements are then used to compute \textit{the force transmission} -- i.e., expected torques around the finger joints ($\tau_{MCP}, \tau_{PIP}$) for a unit actuator force (1 N) through analytical statics equations. Obtaining the force transmission triggers the MATLAB file to evaluate the obtained performance in terms of the constraints and the optimality. Depending on the performance, the optimization algorithm repeats this loop until the requirements are satisfied. 

\subsection{Optimization Algorithms (Evolutionary Algorithms)} \label{sec:optimization_algorithms}

In this paper, we will explore two optimization problems (SOOP and MOOP) using the same optimization algorithms based on EC. EC is a sub-field of soft computing offering nature-inspired algorithms: instead of updating on a single point in the search space, it generates a population of potential solutions (i.e., a set of different combinations of decision variables' values) and evolves them toward the optimum using different metaheuristics (e.g., inspired by genetic recombination, natural selection, universe creation). Since the generic framework of EC methods does not rely on a specific mathematical model and can tackle objective functions defined implicitly through simulations, they are suitable for engineering problems such as optimizing U-HEx.
Specifically, we will employ two different EAs, as described below.

\subsubsection{Genetic Algorithm (GA)} \label{sec:genetic_algorithms}

GAs are the most widely used EAs because they directly replicate the process of natural selection and the concept of survival of the fittest~\cite{goldberg1989genetic,goldberg1990real}. As summarized in Algorithm~\ref{alg:genetic}, GA \textit{(i)} creates a population of random solutions $P$ within the search space, \textit{(ii)} assigns a fitness value to each solution based on the objective function, and
\textit{(iii)} produces new solutions $Q$ by combining the values of the solutions in the current population through a crossover process.
Only the fittest solutions $M$ undergo crossover and are retained in subsequent generations, which allows the GA to evolve its population towards the optimal solution. Similar to biological processes, the crossover operation utilizes the characteristics of high-quality solutions (parents) to generate new, comparable solutions (offspring) -- accelerating convergence to an optimum. However, there is no guarantee that the retrieved optimum is global rather than local. To address this issue, GAs incorporate an additional operator inspired by genetic mutation. This randomly alters the values of newly generated solutions, enhances search-space exploration, and helps in avoiding local optima.

\IncMargin{1em}
\begin{algorithm}[h!]
        \SetKwData{P}{P}
        \SetKwData{X}{M}
        \SetKwData{Q}{Q}
        \SetKwData{G}{g}
        \SetKwData{F}{f}
        \SetKwData{N}{n}
	\SetKwFunction{Init}{randomInitialization}
        \SetKwFunction{Eval}{evaluation}
        \SetKwFunction{Sel}{selection}
        \SetKwFunction{Var}{variation}
        \SetKwFunction{Sur}{survival}
	\SetKwInOut{Input}{input}
	\SetKwInOut{Output}{output}
        \Input{Population size $\N$, number of generations $\G$}
	\Output{The most fitting solution $\P\left(1\right)$}
	\BlankLine		
	
	\Begin{	

    	$\P \leftarrow \Init\left(\N\right)$\;
            $\P \leftarrow \Eval\left(\P\right)$\;
            \For{$i \gets 1 \text{ to } \G$}
            {
                $\X \leftarrow \Sel\left(\P\right)$\;
                $\Q \leftarrow \Var\left(\X\right)$\;
                $\Q \leftarrow \Eval\left(\Q\right)$\;
                $\P \leftarrow \Sur\left(\P,\Q\right)$\;
                
            }
        
		\KwRet{$\P$};
	}	
	\caption{Genetic Algorithm}\label{alg:genetic}
\end{algorithm}\DecMargin{1em}
\subsubsection{Big Bang-Big Crunch Algorithm (BBBC)} \label{sec:bigbang_bigcrunch}

BBBC~\cite{erol2006new} is inspired by the universe's evolution through two phases: explosion and implosion. The process involves energy dissipation, which creates disorder and randomness and then reorganizes this randomness into a new order. As summarized in
Algorithm~\ref{alg:BBBC}, BBBC 
\textit{(i)} generates an initial random population $P$ uniformly distributed across the search space (the explosion, or ``big bang''), and
\textit{(ii)} evaluates and collects the population into their center of mass $cm$ (the implosion, or ``big crunch''). 
As the iterations progress, these two phases are repeated, with new solutions being generated closer to the center of mass, converging to the optimal solution. 
BBBC is recognized for its superior convergence speed compared to GAs, making it a suitable choice for our problem, given the high runtime of the objective function executed in 
a Simulink simulation.

\IncMargin{1em}
\begin{algorithm}
        \SetKwData{P}{P}
        \SetKwData{X}{M}
        \SetKwData{Q}{Q}
        \SetKwData{G}{g}
        \SetKwData{F}{f}
        \SetKwData{N}{n}
        \SetKwData{CM}{cm}
	\SetKwFunction{Init}{randomInitialization}
        \SetKwFunction{Eval}{evaluation}
        \SetKwFunction{Bang}{bang}
        \SetKwFunction{Crunch}{crunch}
        \SetKwFunction{Sel}{selection}
        \SetKwFunction{Var}{variation}
        \SetKwFunction{Sur}{survival}
	\SetKwInOut{Input}{input}
	\SetKwInOut{Output}{output}
        \Input{Population size $\N$, number of generations $\G$}
	\Output{The most fitting solution $\P\left(1\right)$}
	\BlankLine		
	
	\Begin{	

    	$\P \leftarrow \Init\left(\N\right)$\;
            \For{$i \gets 1 \text{ to } \G$}
            {
                \If{$i \neq 1$}{
                    $\P \leftarrow \Bang\left(\CM,i\right)$\;
                }
                $\P \leftarrow \Eval\left(\P\right)$\;
                $\CM \leftarrow \Crunch\left(\P\right)$\;
                
            }
        
		\KwRet{$\P$};
	}
	
	\caption{Big Bang-Big Crunch Algorithm}\label{alg:BBBC}
\end{algorithm}\DecMargin{1em}

\section{Design Process I: Single-Objective Optimization with Mechanical Constraints} \label{sec:design_process_soop}

U-HEx was designed with a linear actuator (Firgelli L12-P) with a maximum displacement of 50 mm \cite{sarac2017design}. In this study, we evaluated the impact of presenting the mechanical limitations of the minimum and maximum linear actuator displacements on U-HEx design by comparing the optimality of their retrieved solutions against the original SOOP~\cite{akbas2024impact}.

\subsection{Optimization Problem (SOOP)} \label{sec:op_1}

Similar to the previous work, the objective function is maximizing the force transmission during the operation: 

\begin{equation}
    \label{eq:obj1}
    \begin{aligned}
    \textrm{maximize}  \quad & \sqrt{\tau_{MCP} \text{ + } \tau_{PIP}} \\
    \end{aligned}
\end{equation}

\noindent where $\tau_{MCP}$ and $\tau_{PIP}$ are calculated based on the values of the decision variables (link lengths). The optimization problem is subject to the following seven inequality constraints: 

\begin{itemize}
    \item $con_{1-4}$: U-HEx is connected to the user’s fingers through passive linear sliders ($c_1$ and $c_2$) whose movements should be limited by the finger size~\cite{sarac2017design}, as 
    Eqn.~(\ref{eq:con_1234});

    \begin{equation}
        \label{eq:con_1234}
        \begin{aligned}
        0 \leq c_1 \leq35; \quad 0 \leq c_2 \leq45
        \end{aligned}
    \end{equation}





    \item $con_{5-6}$: to guarantee a solid grasp and user's safety, the torques $\tau_{MCP}$ and $\tau_{PIP}$ should have a ratio within a reasonable range~\cite{sarac2017design}, as 
    Eqn.~(\ref{eq:con_56})
    ; and

    \begin{equation}
        \label{eq:con_56}
        \begin{aligned}
        \frac{1}{15} \leq \frac{\tau_{MCP}}{\tau_{PIP}} \leq 15
        \end{aligned}
    \end{equation}
    
    \item $con_7$: to ensure a natural range of motion for the hand, the desired actuator displacement ($L_{x}$) must comply with the chosen actuator's mechanical spectrum, as 
    Eqn.~(\ref{eq:con_7}) -- this constraint was not considered in our previous work~\cite{akbas2024impact}. 

    \begin{equation}
        \label{eq:con_7}
        \begin{aligned}
        L_{x}\leq50
        \end{aligned}
    \end{equation}
    
\end{itemize}

Each of these constraints is a function of the decision variables (link lengths) and is calculated in Simulink, similarly to the objective function 
(see Fig.~\ref{fig:schema}(a)~\cite{sarac2017design}). Additional constraints to the problem are related to the decision variable bounds (i.e., the range of link lengths) reported in Table~\ref{tab:bounds}. 

\subsection{Experiment Outline (SOOP)} \label{sec:exp_outline_1}

We designed an experiment with three factors as \textit{(i)} the constraint to the desired actuator displacement (with and without $con_7$), \textit{(ii)} the optimization algorithm (GA and BBBC), and \textit{(iii)} the number of decision variables (six and nine link lengths). When the number of link lengths is limited to six, the additional three links are kept constant as $\overline{BK} = 37$, $\overline{CI} = 16$, and $\overline{EJ} = 37$. These values were selected empirically as detailed in the original work~\cite{sarac2017design}. To ensure a valid comparison, we kept the algorithm parameters the same as our previous work~\cite{akbas2024impact}, reported in Table~\ref{tab:parameter_settings_soop}. Each algorithm was executed 10 times on a computer with 16 core 5.4 GHz CPU and 64 GB RAM. 

\begin{table}[h!]
\footnotesize
    \centering
    \caption{Parameter Settings in SOOP for EAs 
    }
    \begin{tabular}{|c|>{\centering\arraybackslash}c|>{\centering\arraybackslash}c|}
        \hline
        \rowcolor{gray!30} \textbf{PARAMETER} & \textbf{EXPERIMENT VALUE} \\
        \hline
        Max N. of Generations (GA, BBBC) & $50$\\
        \rowcolor{gray!10} Population Size (GA, BBBC) & $300$ \\
        Selection Type (GA) & Binary Tournament~\cite{goldberg1991comparative}\\
        \rowcolor{gray!10} Crossover Type (GA) &  blx-$\alpha$ ($\alpha$ = 0.5)~\cite{eshelman1993real}  \\
        Crossover Probability (GA) & {$1.0$} \\
        \rowcolor{gray!10} Mutation Type (GA) & { Polynomial~\cite{deb1996combined} }\\
        Mutation Probability (GA) & $0.2$ \\
        \rowcolor{gray!10} Survival Strategy (GA) & { Elitist ($\mu + \lambda$ schema)~\cite{beyer2002evolution} } \\
        Crunch Method (BBBC) & { Best Fit~\cite{gencc2010big}}\\
        \rowcolor{gray!10} Constraint Handling (GA, BBBC) & Deb's Method~\cite{coello2002theoretical} \\
        \hline
    \end{tabular}
    \label{tab:parameter_settings_soop}
\end{table}

\subsection{Evaluation Metrics (SOOP)} \label{sec:evaluation_metrics_1}

At each execution, we recorded the set of optimized link lengths (i.e., values of the decision variables belonging to the most fitting solution in the population) and compared each factor in terms of two evaluation metrics.

\subsubsection{Optimality} In SOOP, optimality is the value of the objective function, defined in Eqn.~(\ref{eq:obj1}).
\subsubsection{Convergence Time (CT)} CT indicates the time needed for algorithms to find the optimal solution, defined in Eqn.~(\ref{eq:convergence_time}).

\begin{equation}
    \label{eq:convergence_time}
    \begin{aligned}
        \text{CT} = \frac{\text{GC}}{\text{NG}} \cdot \text{RT}
    \end{aligned}
\end{equation}
\noindent where the overall runtime (RT) indicates the overall time taken by the algorithm and the number of generations (NG) is 50 as indicated in Table~\ref{tab:parameter_settings_soop}. The generation of convergence (GC) is the observed generation in which the algorithm converged to a solution -- specifically if the variance in the value of the retrieved solution did not change in the last 20 iterations with a margin of 0.05 Nm.

\subsection{Results (SOOP)} \label{sec:results_soop}

We analyzed optimality and convergence time using a three-way Repeated Measure Analysis of Variance (RM-ANOVA). Fig.~\ref{fig:exp_results_soop} shows the mean and standard error of the collected data, as reported in Table~\ref{tab:numerical_results_soop}; whereas Table~\ref{tab:anova_soop} summarizes the ANOVA results. 

\subsubsection{Results on Optimality} \label{sec:results_opt_1}

\begin{figure}[b!]
    \centering
    \includegraphics[width=\columnwidth]{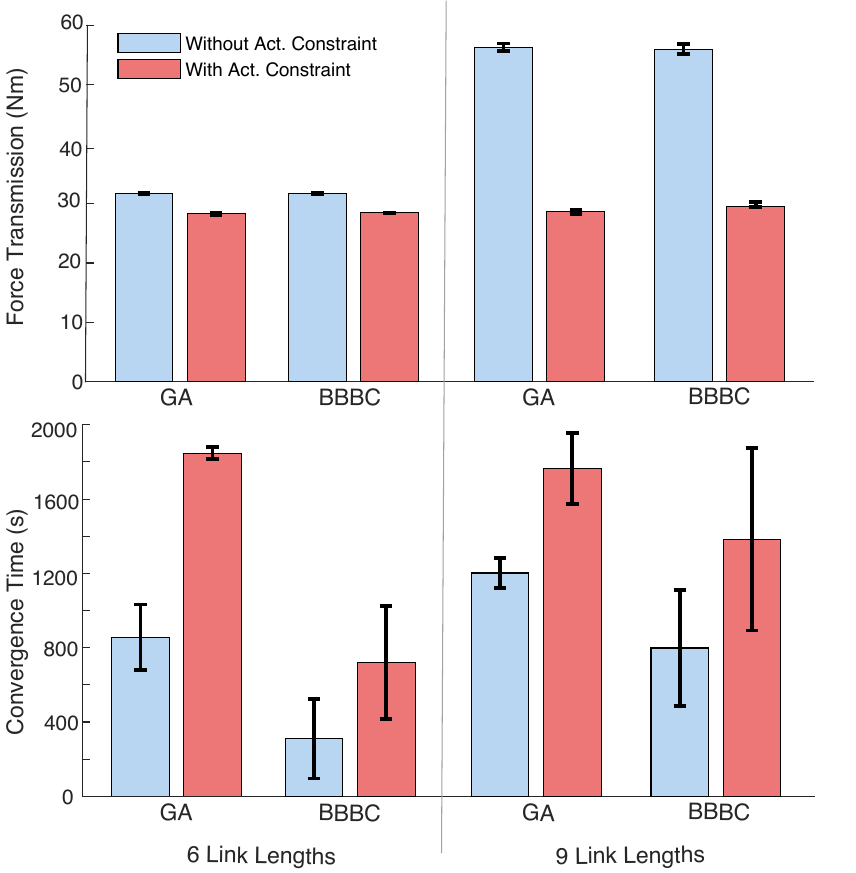}
    \caption{Force transmission (optimality) and convergence time comparisons between experiments (with a different number of variable link lengths): with the constraint to the maximum actuator displacement ($con_7 = L_x \leq 50$) and without~\cite{akbas2024impact}. Plots report the mean and the standard error. 
    }
    \label{fig:exp_results_soop}
\end{figure}

\begin{figure}[h!]
    \centering
    \includegraphics[width=\columnwidth]{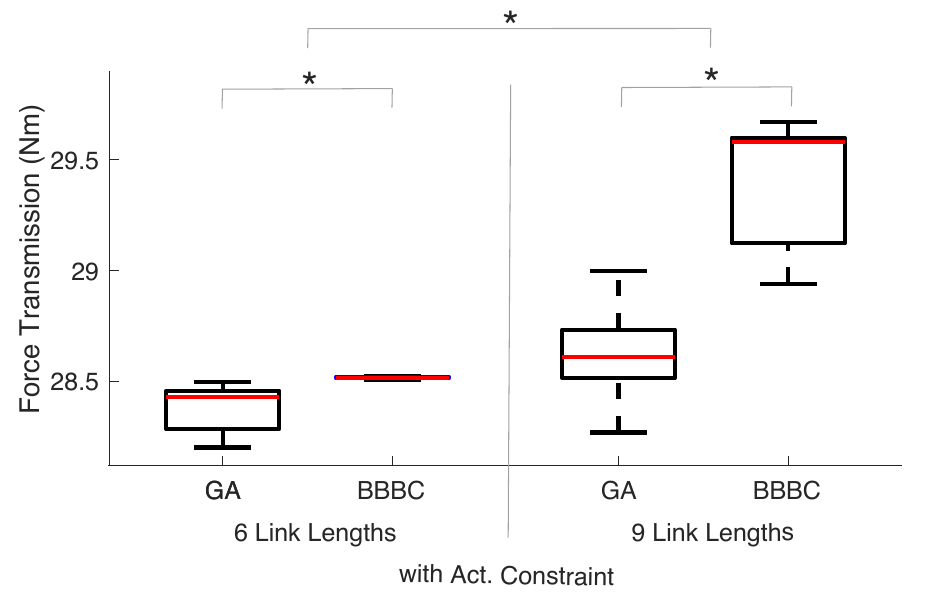}
    \caption{A zoomed-in representation of force transmission (optimality) with the linear actuation constraint ($con_7$) as $L_x \leq 50$. The plots show the median, interquartile range, and outliers of the data. 
    }
    \label{fig:exp_results_soop_zoomed}
\end{figure}

Fig.~\ref{fig:exp_results_soop} shows that implementing the actuation constraint ($con_7$) significantly lowered the force transmission. 
The drop between with and without $con_7$ becomes higher with nine decision variables compared to six. We further analyzed the trends between optimization strategies and decision variables \textit{only} with $con_7$ through a post-hoc analysis, as depicted by a zoomed-in version in Fig.~\ref{fig:exp_results_soop_zoomed}. We found that the optimality results are statistically significantly higher with nine decision variables ($F(1,36)=1047621.937, p<0.001, \eta^2 = 1.000$) and while using BBBC ($F(1,36) = 74.239, p<0.001, \eta^2=0.673$). In addition, we observed their interactions to be statistically significant ($F(1,36) = 38.156, p<0.001, \eta^2=0.515$).

\begin{table}[h!]
\centering
\caption{Numerical Results of Design Process I (SOOP) with 6 and 9 Link Lengths (LL), with and without Actuation Displacement Constraint ($con_7$: $L_x \leq 50$)}
\label{tab:numerical_results_soop}
\resizebox{\columnwidth}{!}{ 
    \begin{tabular}{l c c}
    \multicolumn{3}{c}{\textbf{Force Transmission (Nm)}} \\
    \multicolumn{1}{c}{} & \multicolumn{1}{|c|}{without $con_7$} & with $con_7$ \\ \hline
    GA (6 LL) & \multicolumn{1}{|c|}{$31.63 \pm 0.01$} & \multicolumn{1}{c}{$28.39 \pm 0.10$} \\
    BBBC (6 LL) & \multicolumn{1}{|c|}{$31.65 \pm 0.05$} & \multicolumn{1}{c}{$28.52 \pm 0.005$}  \\ \hline\hline
    GA (9 LL) & \multicolumn{1}{|c|}{$56.20 \pm 0.41$} & \multicolumn{1}{c}{$28.61 \pm 0.19$}  \\
    BBBC (9 LL) & \multicolumn{1}{|c|}{ $55.99 \pm 0.54$} & \multicolumn{1}{c}{$29.43 \pm 0.25$} \\ & & \\
    \multicolumn{3}{c}{\textbf{Run Time (s)}} \\
    \multicolumn{1}{c}{} & \multicolumn{1}{|c|}{without $con_7$} & with $con_7$ \\ \hline
    GA (6 LL) & \multicolumn{1}{|c|}{$860.51 \pm 168.25$} & \multicolumn{1}{c}{$1857.05 \pm 32.84$} \\
    BBBC (6 LL) & \multicolumn{1}{|c|}{$312.50 \pm 204.63$} & \multicolumn{1}{c}{$724.78 \pm 306.87$} \\ \hline\hline
    GA (9 LL) & \multicolumn{1}{|c|}{$1208.59 \pm 78.31 $} & \multicolumn{1}{c}{$1772.48 \pm 182.43$} \\
    BBBC (9 LL) & \multicolumn{1}{|c|}{$802.50 \pm 298.03$} & \multicolumn{1}{c}{$1391.63 \pm 469.05$} \\ 
    \end{tabular}
}
\end{table}


\subsubsection{Results on Convergence Time}  \label{sec:results_convtime_1}

Fig.~\ref{fig:exp_results_soop} shows that adding $con_7$ to the optimization problem significantly increased the convergence time for both EAs and the number of decision variables. We further analyzed the data collected with $con_7$. We further analyzed the trends between optimization strategies and the number of link lengths \textit{only} with $con_7$ through a post-hoc analysis. We found that the convergence time results are statistically significantly different with higher decision variables (($F(1,36) = 9.001, p = 0.005, \eta^2 = 0.200$)) and while using GA instead of BBBC ($F(1,36) = 60.781, p<0.001, \eta^2= 0.628$). In addition, we found their interactions to be statistically significant ($F(1,36) = 14.989, p<0.001, \eta^2= 0.294$). 

\subsection{Discussion (SOOP)} \label{sec:discussion_soop}

We investigated the impact of constraints based on the desired actuator displacement (limitations of the actuator chosen from the market, Firgelli L12-P) on the optimization problem using different EAs. We used two EAs (GA and BBBC) and compared our results to our previous study~\cite{akbas2024impact}.
Our results (Fig.~\ref{fig:exp_results_soop}) show that including the actuation constraint lowers the optimality (i.e., force transmission) while using both optimization methods and decision variable settings (six and nine link lengths), as expected -- therefore, $H_1$ holds true. 

Fig.~\ref{fig:exp_results_soop} also shows that increasing the number of decision variables significantly did not increase the optimality when the actuator constraint was introduced. This might indicate that all the solutions leading to high values of optimality require actuator displacements higher than 50 mm (i.e., unfeasible solutions), which will be investigated in the following design process with MOOP (see Sec.~\ref{sec:design_process_moop}). In addition to a decrease in optimality, we observed that the actuation constraint resulted in statistically higher convergence times for the optimization algorithms. We speculate that higher convergence times might be due to the change in the search space and the computational toll of evaluating one more constraint~\cite{coello2002theoretical}. 

\begin{table}[h!]
\centering
\caption{Statistical Results of Design Process I (SOOP)
} 
\label{tab:anova_soop}
\resizebox{\columnwidth}{!}{ 
    \begin{tabular}{c|c|c}
      &
      Optimality &
      Convergence Time  
    \\ \hline
     {\begin{tabular}[c]{@{}c@{}} Algorithm \\ (GA vs. BBBC) \end{tabular}} &
     \textbf{\begin{tabular}[c]{@{}c@{}}F(1,72)=9.351, \\p$=$0.003,  $\eta^2$= 0.115
     \end{tabular}}
      &
      \textbf{\begin{tabular}[c]{@{}c@{}}F(1,72)=45.624, \\p$<$001,  $\eta^2$= 0.388
       \end{tabular}}
    \\ \hline
     {\begin{tabular}[c]{@{}c@{}} N. of Link Lengths \\ (6 vs. 9) \end{tabular}} &
     \textbf{\begin{tabular}[c]{@{}c@{}}F(1,72)=39221.829, \\p$<$0.001,  $\eta^2$= 0.998
     \end{tabular}}
      &
      \textbf{\begin{tabular}[c]{@{}c@{}}F(1,72)=7.992, \\p$=$0.006,  $\eta^2$= 0.100
       \end{tabular}}
    \\ \hline
     {\begin{tabular}[c]{@{}c@{}} Actuator Constraint \\ (with vs. without) \end{tabular}} &
     \textbf{\begin{tabular}[c]{@{}c@{}}F(1,72)=57325.641, \\p$<$0.001,  $\eta^2$= 0.999
     \end{tabular}}
      &
      \textbf{\begin{tabular}[c]{@{}c@{}}F(1,72)=146.730, \\p$<$0.001,  $\eta^2$= 0.671
       \end{tabular}}
    \\ \hline
    {\begin{tabular}[c]{@{}c@{}} Algorithm $\times$ N. of Link Lengths \end{tabular}} &
    \begin{tabular}[c]{@{}c@{}}F(1,72)=3.209, \\p$=$0.077,  $\eta^2$= 0.043
    \end{tabular}
     &
     \begin{tabular}[c]{@{}c@{}}F(1,72)=0.722, \\p$=$0.398,  $\eta^2$= 0.010
      \end{tabular}
    \\ \hline
    {\begin{tabular}[c]{@{}c@{}} Algorithm $\times$ Constraint \end{tabular}} &
    \textbf{\begin{tabular}[c]{@{}c@{}}F(1,72)=21.140, \\p$<$0.001,  $\eta^2$= 0.227
    \end{tabular}}
     &
     \begin{tabular}[c]{@{}c@{}}F(1,72)=0.164, \\p$=$0.687,  $\eta^2$= 0.002
      \end{tabular}
    \\ \hline
    {\begin{tabular}[c]{@{}c@{}} N. of Link Lengths $\times$ Constraint \end{tabular}} &
    \textbf{\begin{tabular}[c]{@{}c@{}}F(1,72)=35713.120, \\p$<$0.001,  $\eta^2$= 0.998
    \end{tabular}}
     &
     \textbf{\begin{tabular}[c]{@{}c@{}}F(1,72)=11.982, \\p$<$0.001,  $\eta^2$= 0.143
      \end{tabular}}
    \\ \hline
    {\begin{tabular}[c]{@{}c@{}} Algorithm $\times$ \\ N. of Link Lengths $\times$ Constraint \end{tabular}} &
    \textbf{\begin{tabular}[c]{@{}c@{}}F(1,72)=13.668, \\p$<$0.001,  $\eta^2$= 0.160
    \end{tabular}}
     &
     \begin{tabular}[c]{@{}c@{}}F(1,72)=0.046, \\p$=$0.830,  $\eta^2$= 0.001
      \end{tabular}
    \\ \hline
    \end{tabular}
}
\end{table}

We also analyzed the differences between the two EAs and the number of decision variables when the actuation constraint was implemented separately since the optimality results were primarily influenced by the optimization problem without the actuation constraint. Fig.~\ref{fig:exp_results_soop_zoomed} shows that increasing the number of decision variables provides a statistical increase in force transmission -- similar to our previous study~\cite{akbas2024impact}. 

Interestingly, we observed that BBBC offers statistically higher optimality results than GA while working with six and nine decision variables independently -- even though previously we found no difference between them~\cite{akbas2024impact}. We speculate that this was caused by the GA being stuck in a local maxima instead of a global one. 
This is also supported by observing the convergence time: while BBBC reached the current optimal solution\footnote{We remind readers that EC methods retrieve approximate optimal solutions rather than the exact one.} relatively fast, GA continued the search for a significantly longer time. Thus, the convergence time of GA did not change as much as BBBC when the number of decision variables was increased. As mentioned in the state-of-the-art, BBBC was proposed to address some of the biggest disadvantages of GAs, such as premature convergence, convergence speed, and execution time~\cite{erol2006new}. 


\section{Design Process II: Multi-Objective Optimization with Mechanical Constraints} \label{sec:design_process_moop}

The design process in SOOP (Sec.~\ref{sec:design_process_soop}) showed the importance of considering the actuator displacement as a constraint. Introducing a constraint to the desired actuator displacement (Eqn.~\ref{eq:con_7}) reduced the force transmission of the exoskeleton but allowed it to be equipped with smaller actuators that can fit on top of the hand while still exerting a sufficient amount of force to assist users in opening and closing their fingers. 
In the meantime, users' safety was ensured by balancing the torques applied on both finger joints, as detailed in Eqn.~(\ref{eq:con_56}). 

Even though these constraints allowed us to have a design within safe and feasible boundaries, they do not ensure reaching an \textit{optimal} one about these factors. We re-defined the optimization problem to include two further objectives that were previously treated as constraints (i.e., $con_{5-6}$ and $con_7$). We believe that turning constraints into additional objectives can help find better solutions by allowing more flexibility and balancing multiple requirements effectively, resulting in a set of different designs with different properties.

\subsection{Background on Multi-Objective EAs (MOEAs)} \label{sec:background_moop}

MOOP involves optimizing two or more objectives; they usually conflict with each other (i.e., optimizing one might worsen the second), and it is not possible to retrieve a single solution that fully optimizes all the objectives simultaneously. Therefore, optimality in MOOP is defined by the concept of pairwise \textit{dominance}: a solution $x_1$ dominates a solution $x_2$ if $x_1$ is not worse than $x_2$ in all objectives \textit{and} $x_1$ is strictly better than $x_2$ in at least one objective~\cite{deb2001multi}. When solutions do not dominate each other, they are considered as optimal. This results in a set of trade-off solutions (i.e., Pareto Front), each with different degrees of gain and losses to the objectives. 

EAs are great candidates for tackling MOOPs and looking for a set of solutions due to their population-based nature, allowing them to carry on different alternatives at a time~\cite{deb2001multi}. Multi-Objectives Evolutionary Algorithms (MOEAs) aim to retrieve an optimal set (named the Pareto set) with the following properties: \textit{(i)} convergence, or proximity to the true Pareto Front (which is unknown for engineering problems), and \textit{(ii)} diversity, ensuring that the solutions are well-spaced so that the Pareto Front is uniformly covered. These properties are satisfied by different types of \textit{Multi-Objective Survival Strategies} by removing weak solutions from the population. MOEAs usually generate a number of offspring larger than the set population size (to promote space exploration); however, this creates a surplus in the population. Survival strategies deal with this surplus by promoting non-dominated isolated solutions (\textit{elites}) and removing the ones located in crowded regions to promote diversity. For this study, we will explore two of the most well-known strategies in the state-of-the-art.

\paragraph{Non-Dominated Sorting with Crowding Distance (NS)} Introduced with NSGA-II~\cite{deb2002fast}, NS \textit{(i)} divides the population into different sorted non-dominated sets (i.e., the first set dominates all solutions, the second dominates all solutions except the first set, and so on); and \textit{(ii)} deals with the surplus in the population with the crowding distance operator, which sums the normalized distances between neighboring solutions in each objective dimension it identifies isolated solutions (i.e., larger distances indicate more isolated solutions).

\SetCommentSty{mycommfont}
\newcommand\mycommfont[2]{\footnotesize\ttfamily\textcolor{red}{#1}}
\IncMargin{1em}
\begin{algorithm}[h!]
    \SetKwData{P}{P}
    \SetKwData{X}{M}
    \SetKwData{Q}{Q}
    \SetKwData{R}{R}
    \SetKwData{G}{g}
    \SetKwData{F}{f}
    \SetKwData{N}{n}
    \SetKwFunction{Init}{randomInitialization}
    \SetKwFunction{Eval}{evaluation}
    \SetKwFunction{Sel}{selection}
    \SetKwFunction{Var}{variation}
    \SetKwFunction{Sur}{survival}
    \SetKwInOut{Input}{Input}
    \SetKwInOut{Output}{Output}

    \Input{Population size $\N$, number of generations $\G$}
    \Output{The non-dominated set of solutions $\P$}
    \BlankLine

    \Begin{	

    	$\P \leftarrow \Init\left(\N\right)$\;
            $\P \leftarrow \Eval\left(\P\right)$;\hspace{1ex} \tcp{Non-Dominated Sorting + Crowding Distance}
            \For{$i \gets 1 \text{ to } \G$}
            {
                $\X \leftarrow \Sel\left(\P\right)$; \hspace{1ex} \tcp{Crowded Tournament Selection}
                $\Q \leftarrow \Var\left(\X\right)$\;
                $\R \leftarrow \P \cup \Q$\;
                $\R \leftarrow \Eval\left(\R\right)$; \hspace{1ex} \tcp{Non-Dominated Sorting + Crowding Distance}
                $\P \leftarrow \Sur\left(\R\right)$\;
            }
        
		\KwRet{$\P$};
	}	
    \caption{NSGA-II~\cite{deb2002fast}}\label{alg:NSGA-II}
\end{algorithm}
\DecMargin{1em}

\IncMargin{1em}
\begin{algorithm}[h!]
        \SetKwData{P}{P}
        \SetKwData{X}{M}
        \SetKwData{Q}{Q}
        \SetKwData{G}{g}
        \SetKwData{F}{f}
        \SetKwData{N}{n}
	\SetKwFunction{Init}{randomInitialization}
        \SetKwFunction{Eval}{evaluation}
        \SetKwFunction{Str}{strength}
        \SetKwFunction{RF}{rawFitness}
        \SetKwFunction{Dens}{density}
        \SetKwFunction{Sel}{selection}
        \SetKwFunction{Var}{variation}
        \SetKwFunction{Sur}{survival}
	\SetKwInOut{Input}{input}
	\SetKwInOut{Output}{output}
        \Input{Population size $\N$, archive size $\overline{n}$, number of generations $\G$}
	\Output{The non-dominated set of solutions $\overline{P}$}
	\BlankLine		
	
	\Begin{	

    	$\P \leftarrow \Init\left(\N\right)$; \hspace{1ex}\tcp{$\overline{P} = \emptyset$}
            $\P \leftarrow \Eval\left(\P \cup \overline{P}\right)$; \hspace{1ex}\tcp{Fitness = Strength + Density}
            $\overline{P} \leftarrow \Sur\left(\P \cup \overline{P}\right)$;\hspace{1ex}\tcp{$|\overline{P}| = \overline{n}$}
            \For{$i \gets 1 \text{ to } \G$}
            {
                $\X \leftarrow \Sel\left(\overline{P}\right)$\;
                $\P \leftarrow \Var\left(\X\right)$; \hspace{1ex}\tcp{$|\P| = n$}
                $\overline{P} \leftarrow \Eval\left(\P \cup \overline{P}\right)$; \hspace{1ex}\tcp{Fitness = Strength + Density}
            $\overline{P} \leftarrow \Sur\left(\P \cup \overline{P}\right)$;\hspace{1ex}\tcp{Truncation operation, $|\overline{P}| = \overline{n}$}
            }
        
		\KwRet{$\overline{P}$};
	}	
	\caption{SPEA-2~\cite{zitzler2001spea2}}\label{alg:SPEA2}
\end{algorithm}\DecMargin{1em}
\SetCommentSty{mycommfont}
\IncMargin{1em}
\begin{algorithm}
        \SetKwData{P}{P}
        \SetKwData{X}{M}
        \SetKwData{Q}{Q}
        \SetKwData{G}{g}
        \SetKwData{F}{f}
        \SetKwData{N}{n}
        \SetKwData{CM}{cm}
	\SetKwFunction{Init}{randomInitialization}
        \SetKwFunction{Eval}{evaluation}
        \SetKwFunction{Bang}{bang}
        \SetKwFunction{Crunch}{crunch}
        \SetKwFunction{Sel}{selection}
        \SetKwFunction{Var}{variation}
        \SetKwFunction{Sur}{survival}
	\SetKwInOut{Input}{input}
	\SetKwInOut{Output}{output}
        \Input{Population size $\N$, number of generations $\G$}
	\Output{The non-dominated set of solutions $\P$}
	\BlankLine		
	
	\Begin{	

    	$\P \leftarrow \Init\left(\N\right)$\;
            $\P \leftarrow \Eval\left(\P\right)$; \hspace{1ex} \tcp{Non-Dominated Sorting + Crowding Distance}
            \For{$i \gets 1 \text{ to } \G$}
            {
                \If{$i \neq 1$}{
                    $\P \leftarrow \Bang\left(\CM,i,\P\right)$\;
                    $\P \leftarrow \Eval\left(\P\right)$;\hspace{1ex} \tcp{Non-Dominated Sorting + Crowding Distance}
                }
                
                $\CM \leftarrow \Crunch\left(\P\right);$\hspace{1ex} \tcp{$\CM$ is a set of non-dominated solutions}
                
            }
        
		\KwRet{$\P$};
	}
	
	\caption{NS-BBBC}\label{alg:NS-BBBC}
\end{algorithm}\DecMargin{1em}
\IncMargin{1em}
\begin{algorithm}
        \SetKwData{P}{P}
        \SetKwData{X}{M}
        \SetKwData{Q}{Q}
        \SetKwData{G}{g}
        \SetKwData{F}{f}
        \SetKwData{N}{n}
        \SetKwData{CM}{cm}
	\SetKwFunction{Init}{randomInitialization}
        \SetKwFunction{Eval}{evaluation}
        \SetKwFunction{Bang}{bang}
        \SetKwFunction{Crunch}{crunch}
        \SetKwFunction{Sel}{selection}
        \SetKwFunction{Var}{variation}
        \SetKwFunction{Sur}{survival}
	\SetKwInOut{Input}{input}
	\SetKwInOut{Output}{output}
        \Input{Population size $\N$, archive size $\overline{n}$, number of generations $\G$}
	\Output{The non-dominated set of solutions $\overline{P}$}
	\BlankLine		
	
	\Begin{	

    	$\P \leftarrow \Init\left(\N\right)$;  \hspace{1ex}\tcp{$\overline{P} = \emptyset$}
            $\P \leftarrow \Eval\left(\P \cup \overline{P}\right)$; \hspace{1ex}\tcp{Fitness = Strength + Density}
            $\overline{P} \leftarrow \Sur\left(\P \cup \overline{P}\right)$;\hspace{1ex}\tcp{$|\overline{P}| = \overline{n}$}
            \For{$i \gets 1 \text{ to } \G$}
            {
                \If{$i \neq 1$}{
                    $\P \leftarrow \Bang\left(\CM,i,\overline{P}\right)$;  \hspace{1ex}\tcp{$|\P| = n$}
                    $\P \leftarrow \Eval\left(\P \cup \overline{P}\right)$; \hspace{1ex}\tcp{Fitness = Strength + Density}
                $\overline{P} \leftarrow \Sur\left(\P \cup \overline{P}\right)$;\hspace{1ex}\tcp{Truncation operation, $|\overline{P}| = \overline{n}$}
                }
                
                $\CM \leftarrow \Crunch\left(\overline{P}\right)$;\hspace{1ex} \tcp{$\CM$ is a set of non-dominated solutions}
                
            }
        
		\KwRet{$\overline{P}$};
	}
	
	\caption{SP-BBBC}\label{alg:SP-BBBC}
\end{algorithm}\DecMargin{1em}

\paragraph{Strength Pareto Metric with Truncation (SP)} Introduced with SPEA2~\cite{zitzler2001spea2}, SP \textit{(i)} ranks each solution based on how many other solutions it dominates (strength); and \textit{(ii)} deals with the surplus in the population with the truncation operator, which calculates the distance between each solution and its $k$-th closest neighbor, removing the one with smallest distance (i.e., smaller distances indicate more solutions within that radius from the analyzed one).\newline

We applied these two survival strategies to both EAs, resulting in four different MOEAs: the two based on GA are well-known state-of-the-art algorithms (NSGA-II~\cite{deb2002fast} and SPEA2~\cite{zitzler2001spea2}, shown in Algorithms~\ref{alg:NSGA-II} and~\ref{alg:SPEA2}, respectively) whereas the two based on BBBC are new algorithms proposed in this work (NS-BBBC and SP-BBBC, shown in Algorithms~\ref{alg:NS-BBBC} and~\ref{alg:SP-BBBC}, respectively).


\subsection{Optimization Problem (MOOP)} \label{sec:op_2}

The MOOP is composed of three objectives: 

\begin{itemize}
    \item $obj_1$: maximize the force transmission, the same objective optimized in Sec.~\ref{sec:design_process_soop} and expressed in Eqn.~(\ref{eq:obj1}); 
    
    \item $obj_2$: balance the torques around the finger joints for well-distributed forces (i.e., minimize the torque variance), as expressed in Eqn.~(\ref{eq:obj2}) -- minimizing the L1 norm distance between the torque ratio and 1; and

    \begin{equation}
        \label{eq:obj2}
        \begin{aligned}
        \textrm{minimize}  \quad &
        \begin{cases}
          \lvert \frac{\tau_{MCP}}{\tau_{PIP}} - 1 \lvert  & \text{if } \tau_{MCP} \geq \tau_{PIP} \\
          \lvert \frac{\tau_{PIP}}{\tau_{MCP}} - 1 \lvert & \text{otherwise}
        \end{cases}    
        \end{aligned}
    \end{equation}
    
    \item $obj_3$: minimize the required actuator displacement ($L_x$) to assist a natural range of motion for the finger joints, as expressed in Eqn.~(\ref{eq:obj3}).

    \begin{equation}
        \label{eq:obj3}
        \begin{aligned}
        \textrm{minimize}  \quad & L_x
        \end{aligned}
    \end{equation}
    
\end{itemize}

The problem is subject to $con_{1-4}$ defined in Sec.~\ref{sec:op_1} for SOOP, while $con_{5-6}$ and $con_7$ have been turned into $obj_2$ and $obj_3$, respectively. 
Furthermore, we removed $con_7$ to allow the values of $obj_3$ to be unrestricted because we wanted to observe the values of $obj_1$ and $obj_2$ with a hypothetical longer actuator and map their boundaries --  although these solutions will be unfeasible according to our actuation limits.
Additionally, we set the decision variable bounds (i.e., the range of link lengths) 
reported in Table~\ref{tab:bounds}. Since our findings with SOOP showed that using nine decision variables results in better optimality, we used only nine decision variables for the MOOP.

\begin{table}[h!]
    \footnotesize
    \centering
    \caption{Parameter Settings in MOOP for MOEAs
    }
    \begin{tabular}{|c|>{\centering\arraybackslash}c|>{\centering\arraybackslash}c|}
        \hline
        \rowcolor{gray!30} \textbf{PARAMETER} & \textbf{EXPERIMENT VALUE} \\
        \hline
        Max N. of Generations (GA, BBBC) & $100$\\
        \rowcolor{gray!10} Population Size (GA, BBBC) & $300$ \\
        Selection Type (GA) & Binary Tournament~\cite{goldberg1991comparative}\\
        \rowcolor{gray!10} Crossover Type (GA) &  blx-$\alpha$ ($\alpha$ = 0.5)~\cite{eshelman1993real}  \\
        Crossover Probability (GA) & {$1.0$} \\
        \rowcolor{gray!10} Mutation Type (GA) & { Polynomial~\cite{deb1996combined} }\\
        Mutation Probability (GA) & $0.2$ \\
        \rowcolor{gray!10} Survival Strategy (GA, BBBC) & NS~\cite{deb2002fast}, SP~\cite{zitzler2001spea2} \\
        Crunch Method (BBBC) & { Best Fit~\cite{gencc2010big}}\\
        \rowcolor{gray!10} Constraint Handling (GA, BBBC) & Constrained Tournament~\cite{deb2001multi} \\
        \hline
    \end{tabular}
    \label{tab:parameter_settings_moop}
\end{table}

\subsection{Experiment Outline (MOOP)} \label{sec:exp_outline_2}

We designed an experiment with two factors as \textit{(i)} the optimization algorithm (GA and BBBC) and \textit{(ii)} the elitist multi-objective survival strategy (NS and SP). 
The algorithm parameters are reported in Table~\ref{tab:parameter_settings_moop} and retain the same values in Table~\ref{tab:parameter_settings_soop} for the operation shared between SOOP and MOOP (e.g., crossover, mutation, bang, crunch). We increased the maximum number of generations from 50 to 100 because multi-objective algorithms tend to converge slower compared to single-objective due to the increased complexity of optimizing multiple conflicting objectives and computational effort required to maintain a diverse set of solutions and evaluate Pareto dominance~\cite{deb2002fast}. 
Each algorithm was executed 10 times on a computer with 16 core 5.4 GHz CPU and 64 GB RAM. 

\begin{figure}[b!]
    \centering
    \includegraphics[width=\columnwidth]{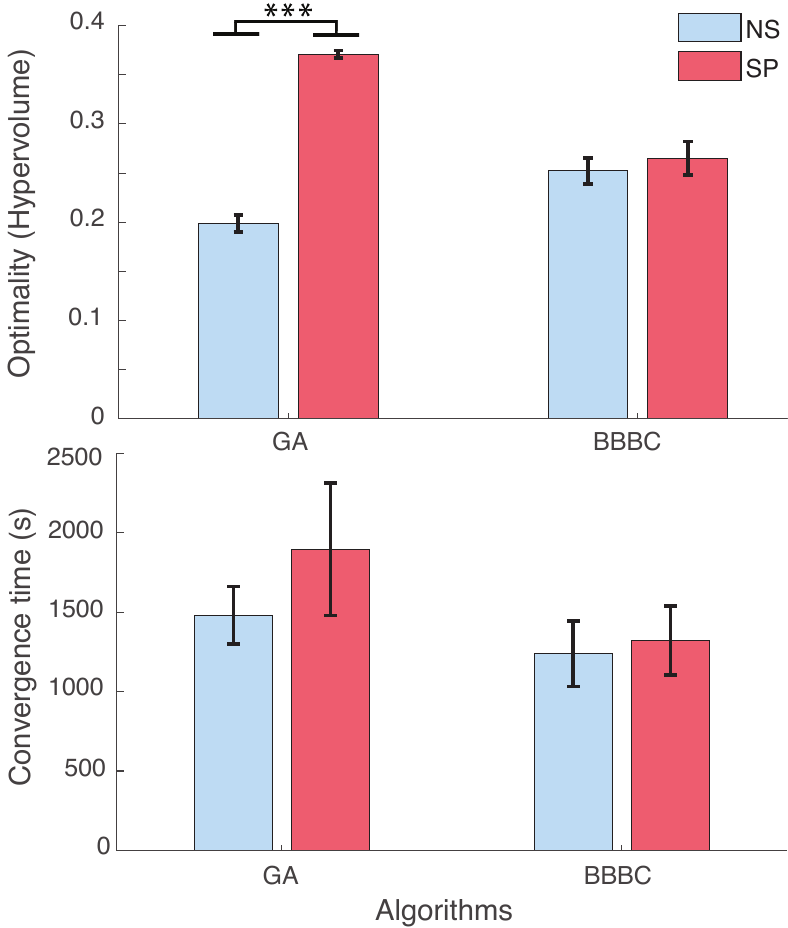}
    \caption{Hypervolume indicator (optimality) and convergence time comparisons between MOOP experiments. Plots report the mean and the standard error. }
    \label{fig:exp_results_moop}
\end{figure}

\subsection{Evaluation Metrics (MOOP)} \label{sec:evaluation_metrics_2}

At each execution, we recorded the set of optimized link lengths (i.e., values of the decision variables belonging to the most fitting non-dominated set) and compared different runs with different factors in terms of two evaluation metrics. 

\subsubsection{Optimality} The optimality of MOOPs is based on two factors: \textit{(i)} how close the solutions are to the Pareto front (convergence); and \textit{(ii)} how diverse they are from each other in the objective space (diversity), such that the Pareto front is uniformly discovered. We evaluated these factors with the \textit{hypervolume indicator}~\cite{shang2020survey}, the volume measurement of the objective space covered by a set of solutions normalized over the objectives (thus, unitless within $[0,1]$). For problems with more than two objectives, the hypervolume indicator is subject to approximation instead of an exact geometrical value~\cite{fonseca2006improved,while2006faster}. Still, it summarizes the convergence to the Pareto Front and the diversity of the solutions in the retrieved set in a single value: larger values indicate better performance. 

\subsubsection{Convergence Time (CT)} Convergence time is calculated as in Eqn.~(\ref{eq:convergence_time}) similar to SOOP (Sec.~\ref{sec:op_1}), except GC is calculated on the hypervolume on a margin of $\expnumber{1.0}{-4}$.

\begin{figure*}[h!]
    \centering
    \subfigure[\protect\url{}\label{fig:pareto_all} All objectives]
    {\includegraphics[width=\columnwidth]{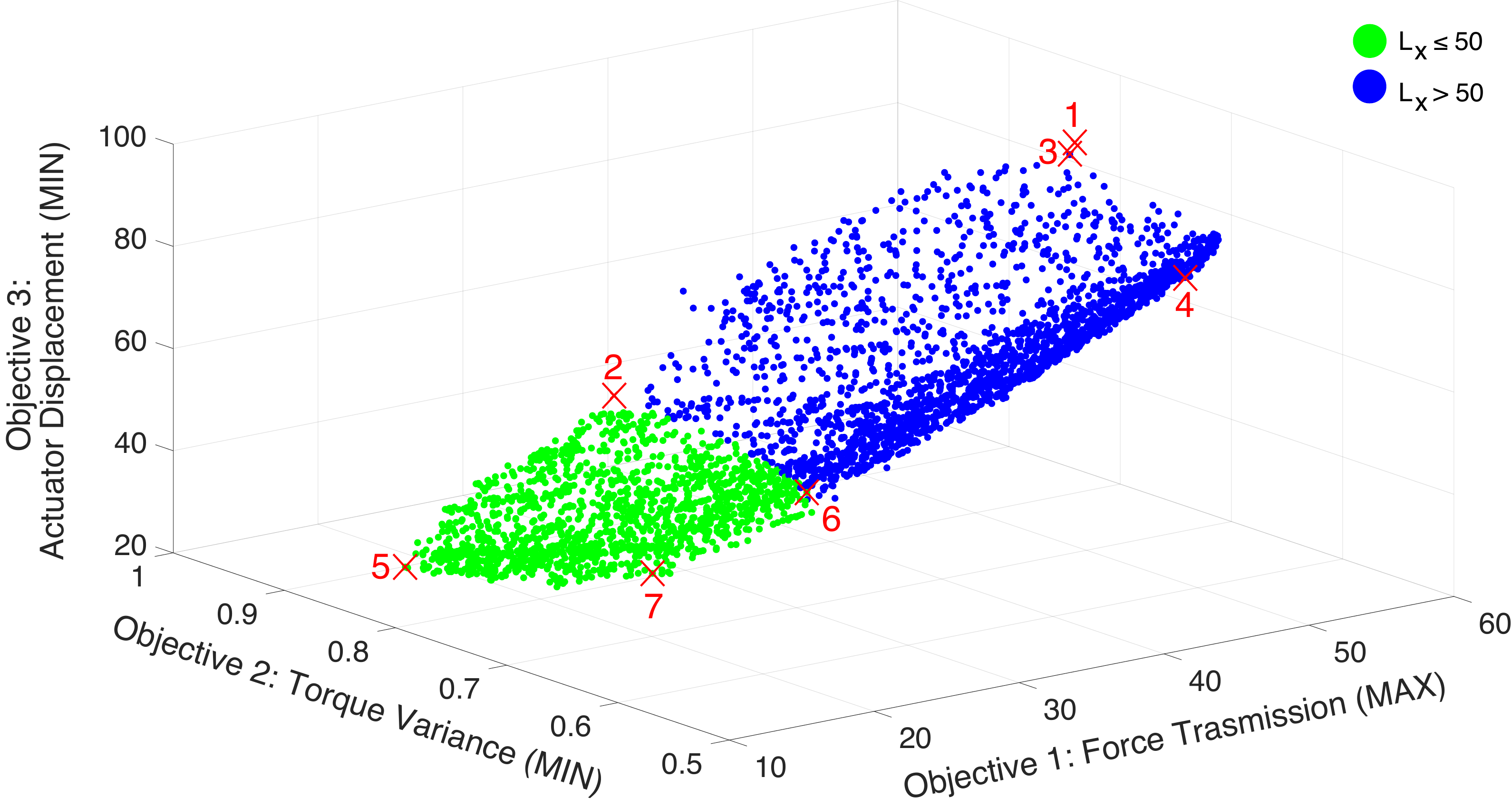}} \hfill
    \subfigure[\protect\url{}\label{fig:pareto_12}Projection $obj_1-obj_2$]
    {\includegraphics[width=\columnwidth]{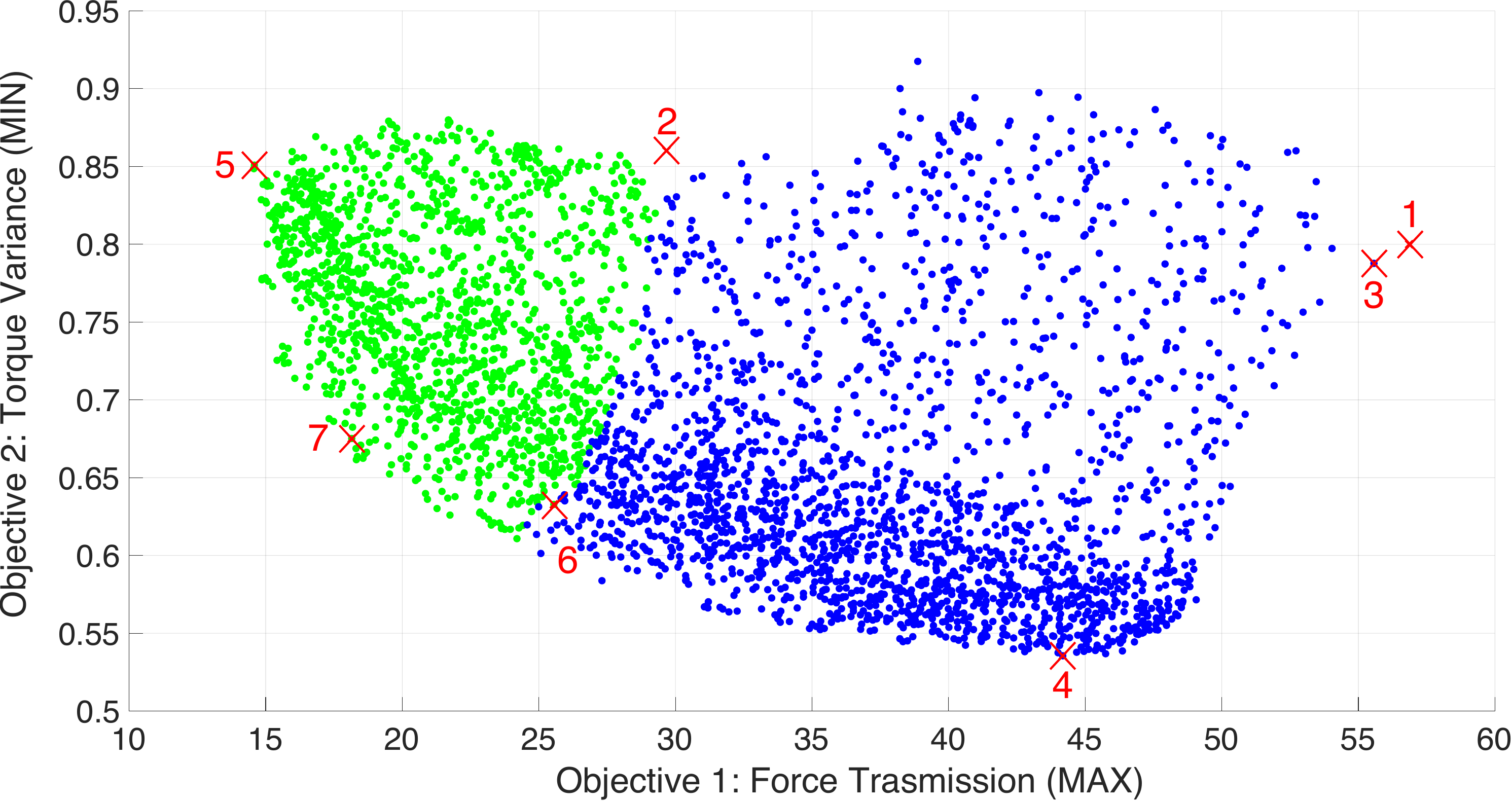}}

    \subfigure[\protect\url{}\label{fig:pareto_13}Projection $obj_1-obj_3$]
    {\includegraphics[width=\columnwidth]{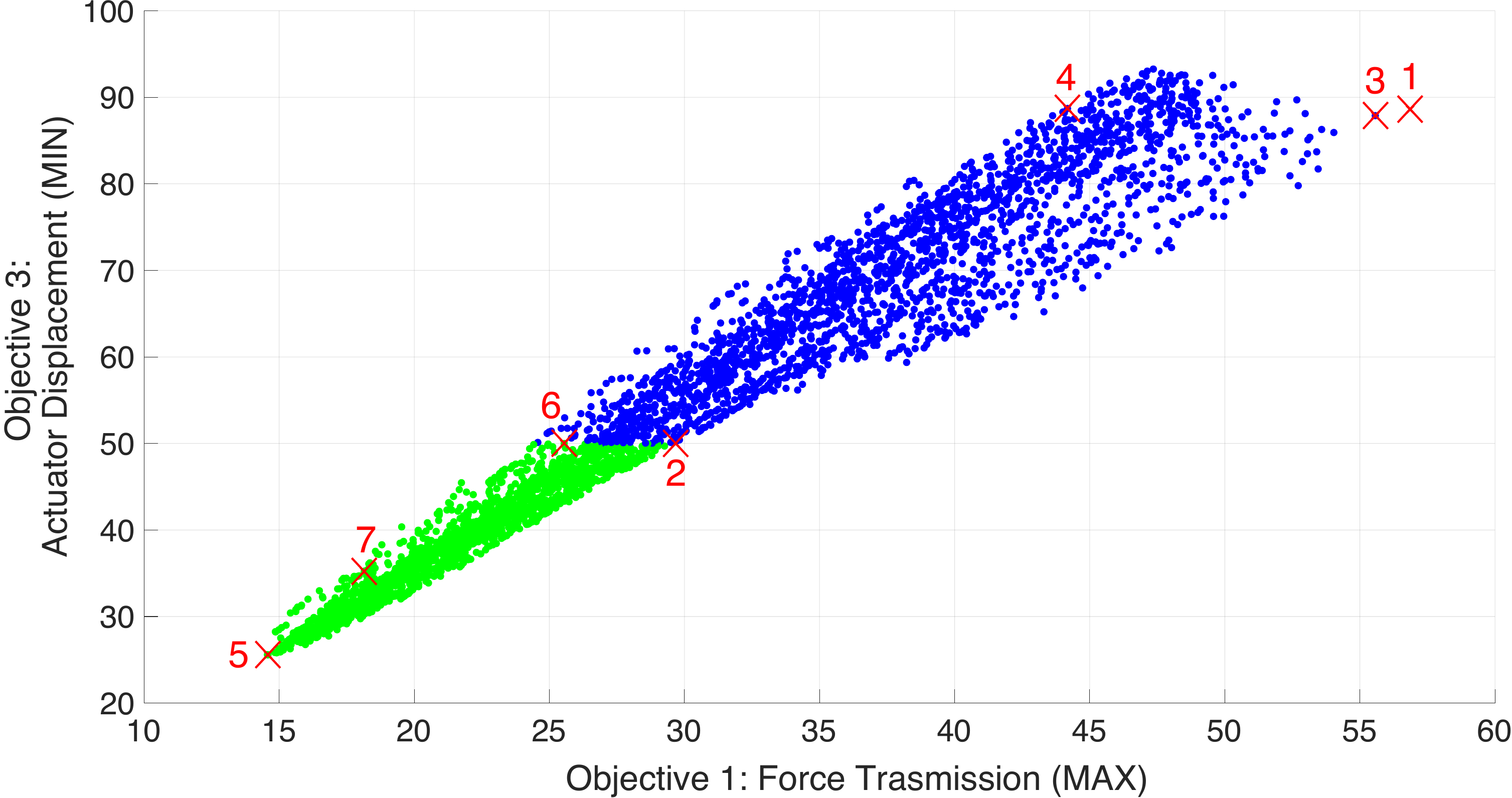}} \hfill
    \subfigure[\protect\url{}\label{fig:pareto_23}Projection $obj_2-obj_3$]
    {\includegraphics[width=\columnwidth]{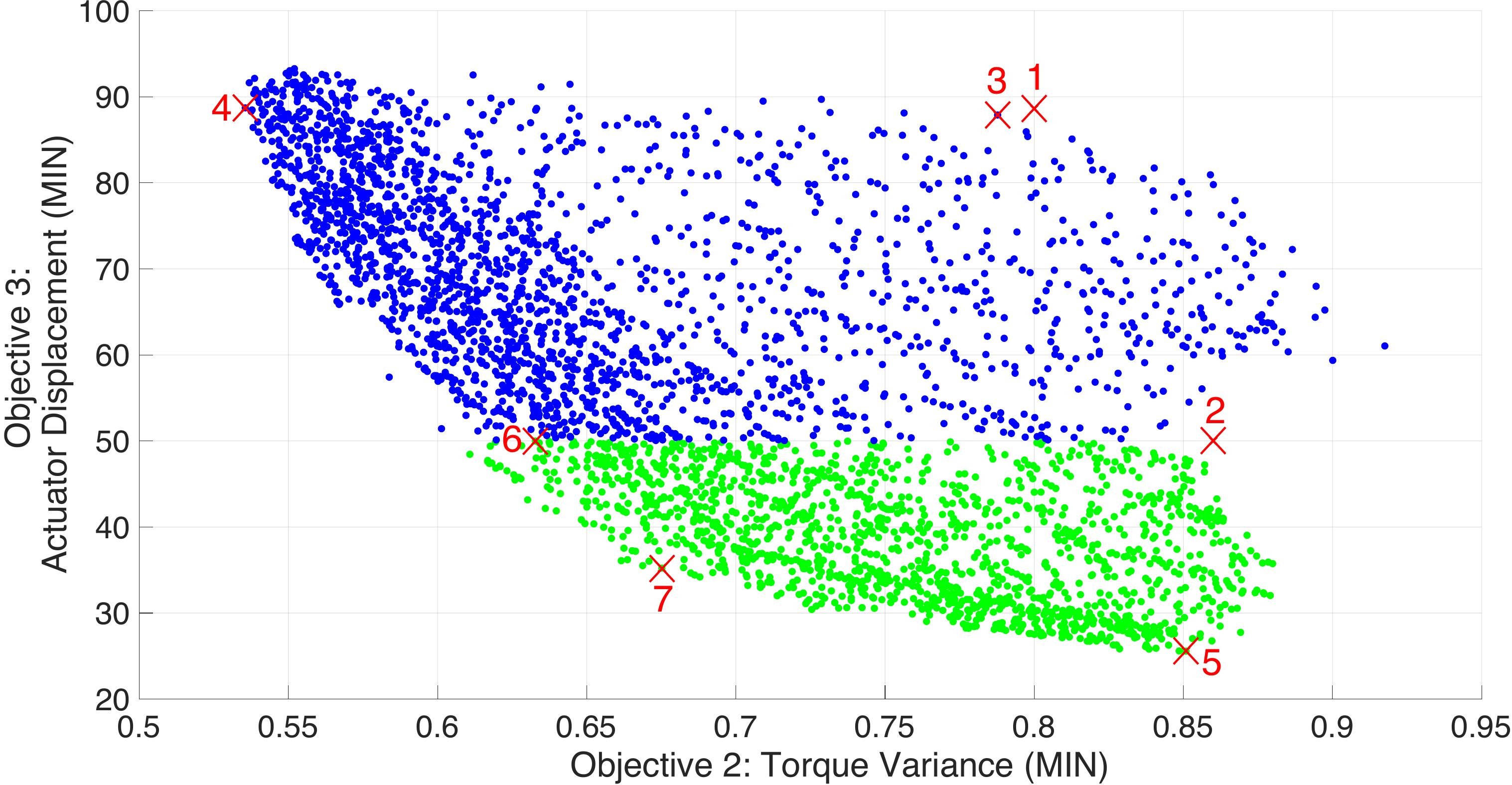}}
    \caption{All non-dominated solutions (i.e., designs) retrieved by four algorithms, collecting 3665 trade-off designs. Solutions with $L_x \leq 50$ are highlighted in green. Designs 1 and 2 are from the SOOP (reported in the figure but not belonging to this set), while Designs 3-7 are five selected designs (reported in Table~\ref{tab:best_designs_values}). Projection (c) shows a linear relation between force transmission and actuator displacement: longer actuators allow for higher force transmission.}
    \label{fig:pareto}
\end{figure*}

\subsection{Results (MOOP)} \label{sec:results_2}

For each MOEA, we analyzed optimality and convergence time using a two-way RM-ANOVA. Fig.~\ref{fig:exp_results_moop} shows the mean and standard error of the collected data, whereas Table~\ref{tab:numerical_results_moop} reports the numerical results.

\subsubsection{Results on Optimality} \label{sec:results_opt_2}

Each algorithm was executed 10 times, each leading to 300 solutions. In total, out of the ($4 \cdot 10 \cdot 300 =$) $12000$ solutions, we found 3665 non-dominated solutions representing different trade-off exoskeleton designs as shown in  Fig.~\ref{fig:pareto}. Each point represents the value in objective space (three dimensions) of a different combination of link lengths in the decision variable space (nine dimensions). The excluded $8335$ dominated solutions are worse than some of the non-dominated solutions in all the objectives.

Fig.~\ref{fig:pareto_13} indicates a consistent and linear relationship between the objective functions related to the force transmission ($obj_1$) and to the actuator displacement ($obj_3$), making the Pareto front \textit{almost} lie on a plane. Thus, minimizing the actuator displacement results in lower force transmission while maximizing the force transmission requires larger amounts of actuator displacements -- in line with what we found in the SOOP design (Sec.~\ref{sec:results_soop}). We quantified the strength of this linear relationship with the \texttt{corrcoef} function from MATLAB, which yielded a coefficient of $0.976$ (with $1$ being the maximum strength), showing a high correlation. However, we did not observe any correlation between the other two objective functions (i.e., force transmission and torque variance or actuator displacement and torque variance).

\begin{table}[h!]
\centering
\caption{Numerical Results of Design Process II (MOOP)}
\label{tab:numerical_results_moop}
\resizebox{\columnwidth}{!}{ 
    \begin{tabular}{l c c || c c}
    &\multicolumn{2}{c}{\textbf{Hypervolume Indicator}} & \multicolumn{2}{c}{\textbf{Run Time (s)}}  \\
    \multicolumn{1}{c}{} & \multicolumn{1}{|c|}{NS} & SP  & \multicolumn{1}{|c|}{NS} & SP \\ \hline
    GA & \multicolumn{1}{|c|}{$0.19 \pm 0.02$} & $0.37 \pm 0.01$ & \multicolumn{1}{|c|}{$1479 \pm 574$} & $1894 \pm 1315 $\\
    BBBC & \multicolumn{1}{|c|}{$0.25 \pm 0.04$} & $0.26 \pm 0.05$ & \multicolumn{1}{|c|}{$1237 \pm 652$} & $1320 \pm 683$ \\ \\
    \end{tabular}
}
\end{table}

Our two-way ANOVA results (Table~\ref{tab:anova_moop}) indicate that the optimality obtained by GA is not significantly different than BBBC. Yet, SP was found to be significantly better than NS -- as it covered a wider area of the Pareto Front (see Table~\ref{tab:numerical_results_moop}). We also observed that the interactions were significantly different than each other: NS was found to be more effective when used for BBBC, while SP was found to be more effective when used for GA. In addition, we observe that choosing the survival strategy critically changed the optimality while using GA ($p<0.001$) while it did not create such a difference while using BBBC.


\subsubsection{Results on Convergence Time}  \label{sec:results_convtime_2}

Fig.~\ref{fig:exp_results_moop} and Table~\ref{tab:numerical_results_moop} show the convergence time results obtained for all algorithms categorized as EAs and survival strategies. Table~\ref{tab:anova_moop} indicates that there is no statistical difference between EAs, survival strategies, nor their interaction.

\begin{table}[h!]
\centering
\caption{Statistical Results of Design Process II (MOOP)
}
\label{tab:anova_moop}
\resizebox{\columnwidth}{!}{ 
    \footnotesize
    \begin{tabular}{c|c|c}
      &
      Optimality &
      Convergence Time  
    \\ \hline
     {\begin{tabular}[c]{@{}c@{}} Algorithm \\ (GA vs. BBBC) \end{tabular}} &
     \begin{tabular}[c]{@{}c@{}}F(1,36)=3.159, \\ p$=$0.109,  $\eta^2$= 0.260
     \end{tabular}
      &
      \begin{tabular}[c]{@{}c@{}}F(1,36)= 2.25, \\p$=$0.1421,  $\eta^2$= 0.059
       \end{tabular}
    \\ \hline
     {\begin{tabular}[c]{@{}c@{}} Survival Strategy \\ (NS vs SP) \end{tabular}} &
     \textbf{\begin{tabular}[c]{@{}c@{}}F(1,36)=62.742, \\p$<$0.001,  $\eta^2$= 0.875
     \end{tabular}}
      &
      \begin{tabular}[c]{@{}c@{}}F(1,36)=0.84, \\p$=$0.3658,  $\eta^2$= 0.023
       \end{tabular}
    \\ \hline
     {\begin{tabular}[c]{@{}c@{}} Algorithm $\times$ \\ Survival Strategy \end{tabular}} &
     \textbf{\begin{tabular}[c]{@{}c@{}}F(1,36)=51.338, \\p$<$0.001,  $\eta^2$= 0.851
     \end{tabular}}
      &
      \begin{tabular}[c]{@{}c@{}}F(1,36)=0.37, \\p$=$0.5446,  $\eta^2$= 0.010
       \end{tabular}
    \\ \hline
    \end{tabular}
}
\end{table}

\subsection{Discussion (MOOP)} \label{sec:discussion_2}

We observed the impact of optimizing multiple objective functions simultaneously while using four different MOEAs. MOEAs' efficiency in retrieving optima depends on survival strategies, which promote diversity by emphasizing non-dominated isolated solutions and removing the ones located in crowded regions. We tested two strategies (NS and SP) on two EAs (GA and BBBC) by redefining the SOOP by turning previous constraints into new objective functions. 


In terms of optimality, we found no significant difference between different EAs (see Fig.~\ref{fig:exp_results_moop} and Table~\ref{tab:anova_moop}) -- similar to SOOP without $con_7$~\cite{akbas2024impact} but not to SOOP with $con_7$. Previously, in Sec.~\ref{sec:discussion_soop}, we reported that BBBC showed significantly better optimality with $con_7$ and speculated that it might be caused by GA getting stuck in a local optimum rather than a global one. The MOOP results support this speculation since unconstraining the actuator displacement yielded similar trends to the results of SOOP without $con_7$ -- despite having a more complex optimization problem. 

We also observed that changing the survival strategy significantly changed the optimality -- depending on the EA. Overall, using SP as a survival strategy is significantly better than NS.  
Furthermore, when applied to GA, NS was observed to be significantly better than SP, which is in line with SPEA2 providing more optimal and diverse solutions at the expense of computational time~\cite{king2010comparison}. Even though SP-BBBC features a more uniform distribution than the one retrieved by NS-BBBC, we did not observe the same statistical difference with BBBC. 

Overall, even though BBBC performs not statistically differently than GA, it yields more consistent results while changing conditions in the optimization problem (e.g., adding constraints and/or using another survival strategy). Both optimization methods are valuable for different audiences: designers with a strong background in optimization (who might need many different conditions) could obtain a better set of solutions with GA at the cost of more intense preliminary work. Yet, designers who are unfamiliar with optimization might obtain more consistent results using BBBC.

In addition, designers must be mindful of the objective functions before performing the experiment. Fig.~\ref{fig:pareto_13} shows a strict correlation between force transmission and desired actuator displacement: increasing the actuator displacement results in higher force transmission. Therefore, maximizing the transmission while minimizing the displacement simultaneously would be an impossible task -- a perfect example of conflicting objectives. To avoid possible unfeasible solutions and computational burden, designers might choose to treat one of these objectives as a constraint. 
If the actuator allowed for higher displacements, having both objectives in the MOOP might be useful to explore alternative solutions -- at the cost of increased weight for the exoskeleton, which must support a bigger and possibly more cumbersome actuator. 

Lastly, 
treating constraints as further objectives offered distinct advantages by aiming to find solutions closest to a theoretical \textit{ideal} point, where each objective achieves its optimal value. The set depicted in Fig.~\ref{fig:pareto}, which is a combination of solutions retrieved by four MOEAs over 40 runs, is a comprehensive evaluation of trade-offs featuring optimal solutions that are not constrained. Ultimately, this allowed us to trace the shape of the objective space for U-HEx, understand the relationships between the objectives, and identify regions of interest. Therefore, $H_2$ holds true.

\begin{table*}[h!]
\centering
\caption{Link Lengths and Objective Values of the Selected Best Designs (9 Link Lengths). \textbf{Bold} Objective Values indicate Design Selection Criteria. \todo{Red} Obj3 Values indicate Unfeasible Solutions due to Actuator Displacement longer than 50 mm}
\label{tab:best_designs_values}
\resizebox{\textwidth}{!}{%
\begin{tabular}{
>{\columncolor[HTML]{FFFFFF}}l |
>{\columncolor[HTML]{FFFFFF}}c |
>{\columncolor[HTML]{FFFFFF}}c |
>{\columncolor[HTML]{FFFFFF}}c ||
>{\columncolor[HTML]{FFFFFF}}c |
>{\columncolor[HTML]{FFFFFF}}c |
>{\columncolor[HTML]{FFFFFF}}c |
>{\columncolor[HTML]{FFFFFF}}c |
>{\columncolor[HTML]{FFFFFF}}c |
>{\columncolor[HTML]{FFFFFF}}c |
>{\columncolor[HTML]{FFFFFF}}c |
>{\columncolor[HTML]{FFFFFF}}c |
>{\columncolor[HTML]{FFFFFF}}c |
>{\columncolor[HTML]{FFFFFF}}c |
>{\columncolor[HTML]{FFFFFF}}c |
>{\columncolor[HTML]{FFFFFF}}c |}
\cline{2-16}
 & ${obj_1}$ & ${obj_2}$ & ${obj_3}$ & 
 ${\overline{BC}}$ &
 ${\overline{BK}}$ & ${\overline{EF}}$ & ${\overline{DE}}$ & ${\overline{CD}}$ & ${\overline{EJ}}$ & ${\overline{LLX}}$ & ${\overline{CI}}$ & ${\overline{AB}}$ & ${\overline{KH}}$ & ${\overline{GH}}$ & ${\overline{GF}}$  \\ \hline
\multicolumn{1}{|l|}{{Design 1 (SOOP)}} & \textbf{56.86} & 0.80 & \todo{88.60} & 60.00 & 48.50 & 51.00 & 15.00 & 10.00 & 36.54 & 20.00 & 10.98 & 20.00 & 72.00 & 91.37 & 56.00  \\ \hline
\multicolumn{1}{|l|}{{Design 2 (SOOP)}} & \textbf{29.67} & 0.86 & 50.00 & 57.09 & 40.30 & 15.00 & 15.00 & 10.00 & 42.72 & 20.00 & 16.00 & 20.00 & 72.00 & 100.00 & 56.00  \\ \hline
\multicolumn{1}{|l|}{{Design 3 (MOOP)}} & \textbf{55.57} & 0.79 & \todo{87.88}  & 60.00 & 47.78 & 50.87 & 15.05 & 10.13 & 36.07 & 20.00 & 10.00 & 20.00 & 72.00 & 91.73 & 56.00 \\ \hline
\multicolumn{1}{|l|}{{Design 4 (MOOP)}} & 44.18 & \textbf{0.54} & \todo{88.70} & 59.54 & 46.65 & 50.54 & 15.01 & 16.81 & 50.00 & 20.00 & 10.29 & 20.00 & 72.00 & 98.15 & 56.00  \\ \hline
\multicolumn{1}{|l|}{{Design 5 (MOOP)}}& 14.58 & 0.85 & \textbf{25.57} & 38.07 & 20.00 & 15.99 & 29.51 & 12.65 & 39.84 & 20.00 & 10.37 & 20.00 & 72.00 & 92.46 & 31.36  \\ \hline
\multicolumn{1}{|l|}{{Design 6 (MOOP)}}& \textbf{25.55} & 0.63 & 49.99 & 46.59 & 26.30 & 51.00 & 21.25 & 12.51 & 50.00 & 20.00 & 10.83 & 20.00 & 72.00 & 88.97 & 55.97  \\ \hline
\multicolumn{1}{|l|}{{Design 7 (MOOP)}} & \textbf{18.15} & \textbf{0.68} & \textbf{35.22} & 38.00 & 20.00 & 44.84 & 34.81 & 11.14 & 48.95 & 20.00 & 10.96 & 20.00 & 72.00 & 94.62 & 55.43 \\ \hline
\end{tabular}%
}
\end{table*}

\begin{figure*}[h!]
    \centering
    \subfigure[\protect\url{}\label{fig:cad2}Design 2 (SOOP)]
    {\includegraphics[height=4.5cm]{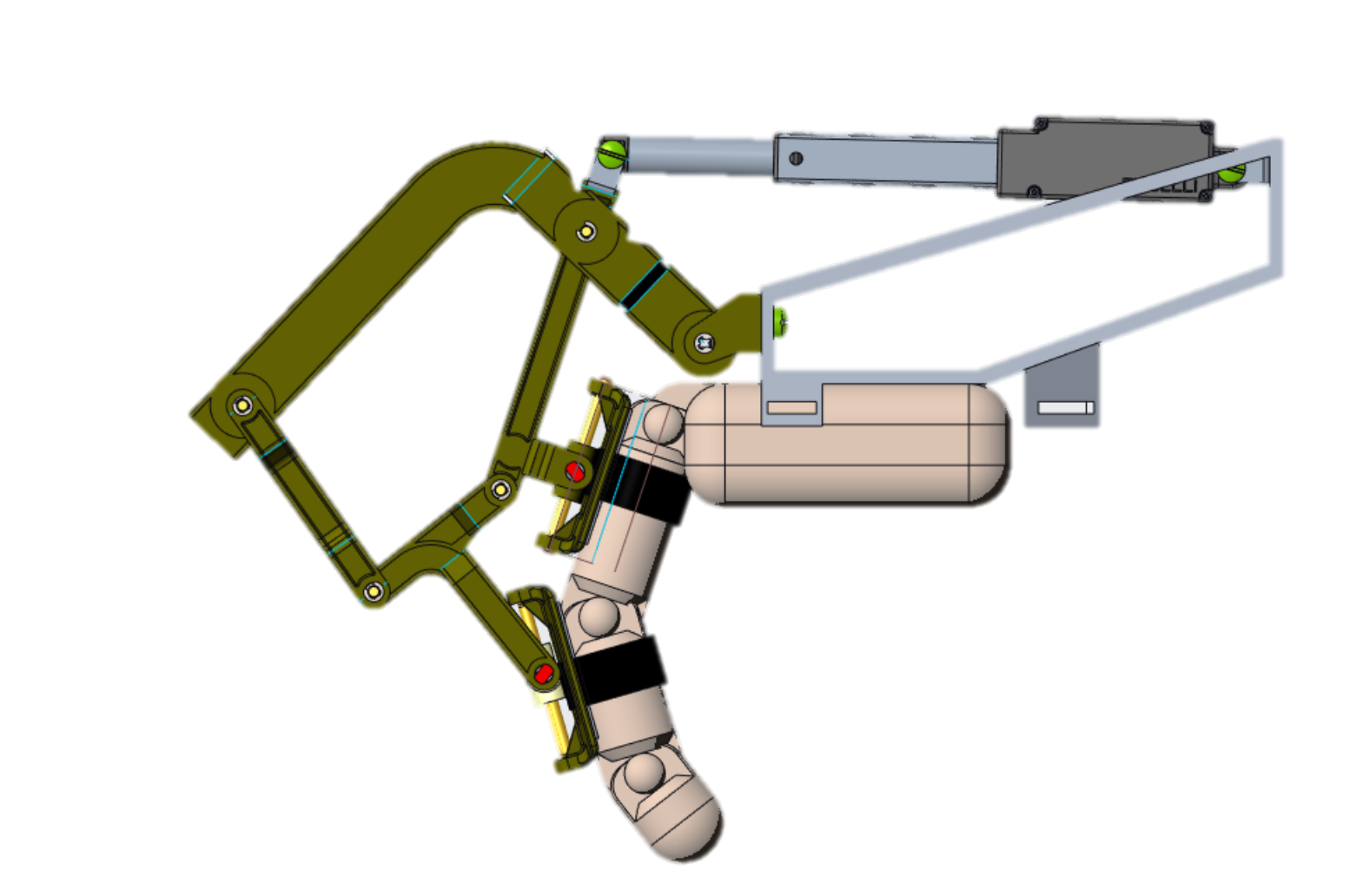}}
    \subfigure[\protect\url{}\label{fig:cad5}Design 5 (MOOP)]
    {\includegraphics[height=4.5cm]{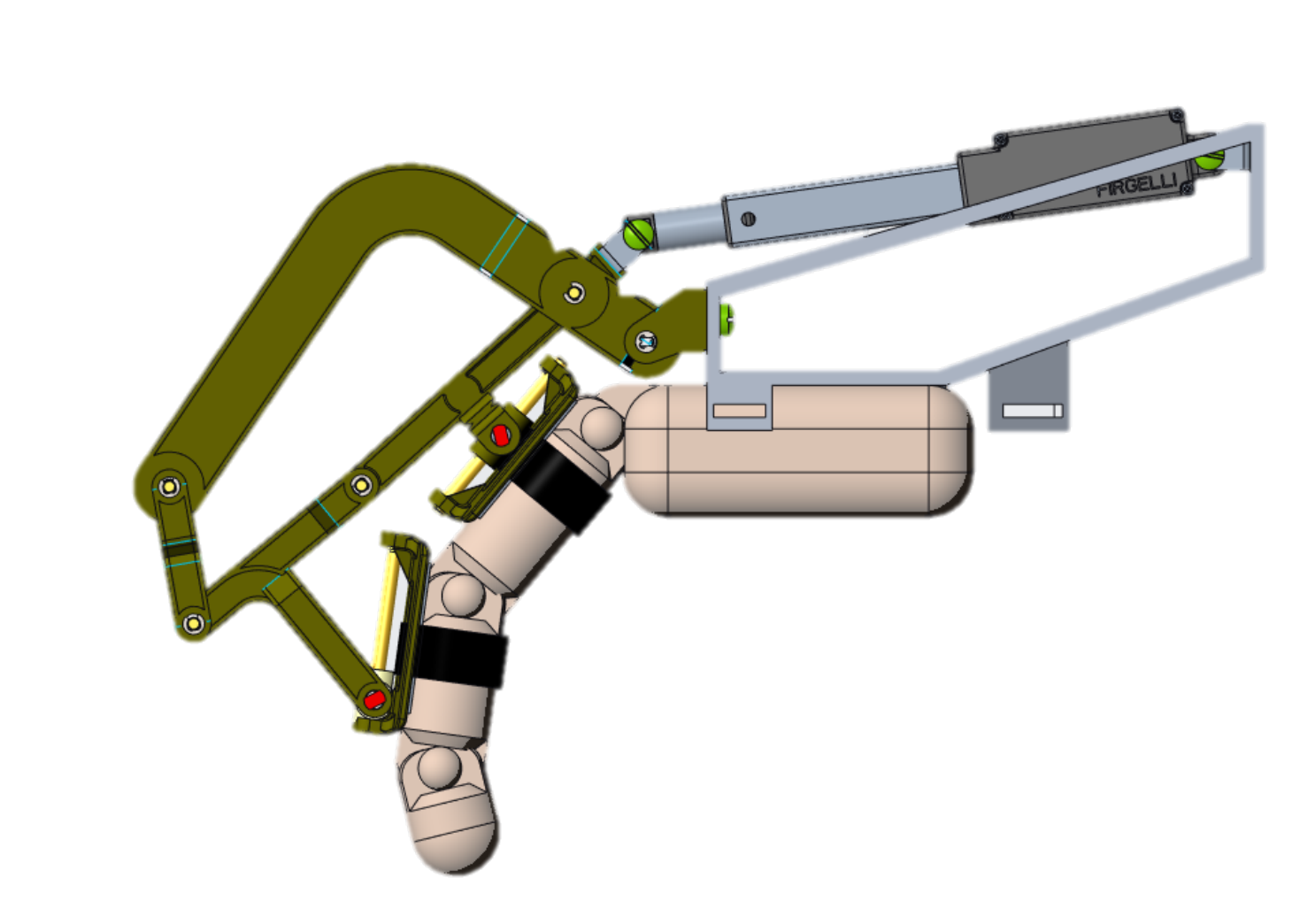}}
    \subfigure[\protect\url{}\label{fig:cad6}Design 6 (MOOP)]
    {\includegraphics[height=4.5cm]{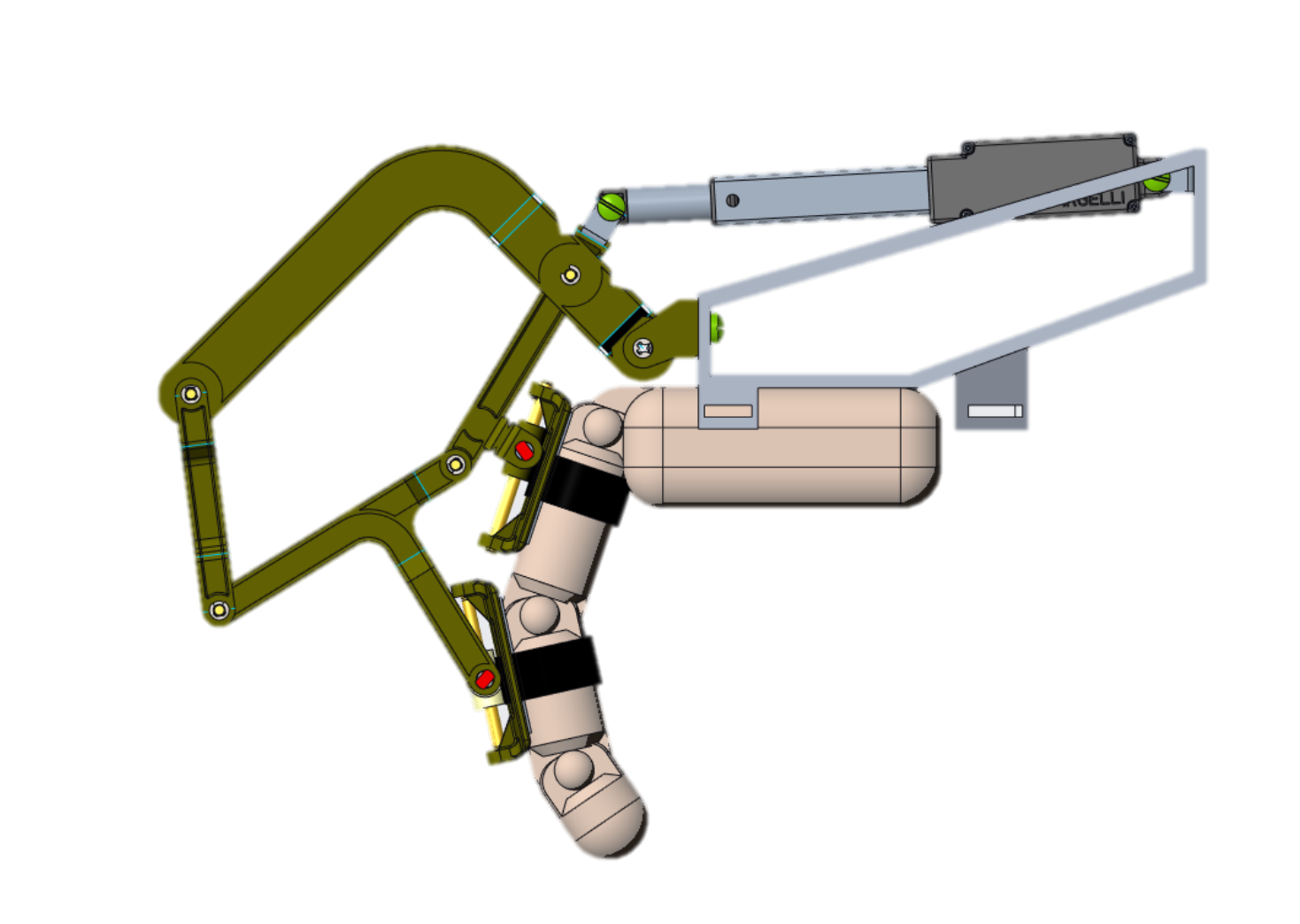}}
    \subfigure[\protect\url{}\label{fig:cad7}Design 7 (MOOP)]
    {\includegraphics[height=4.5cm]{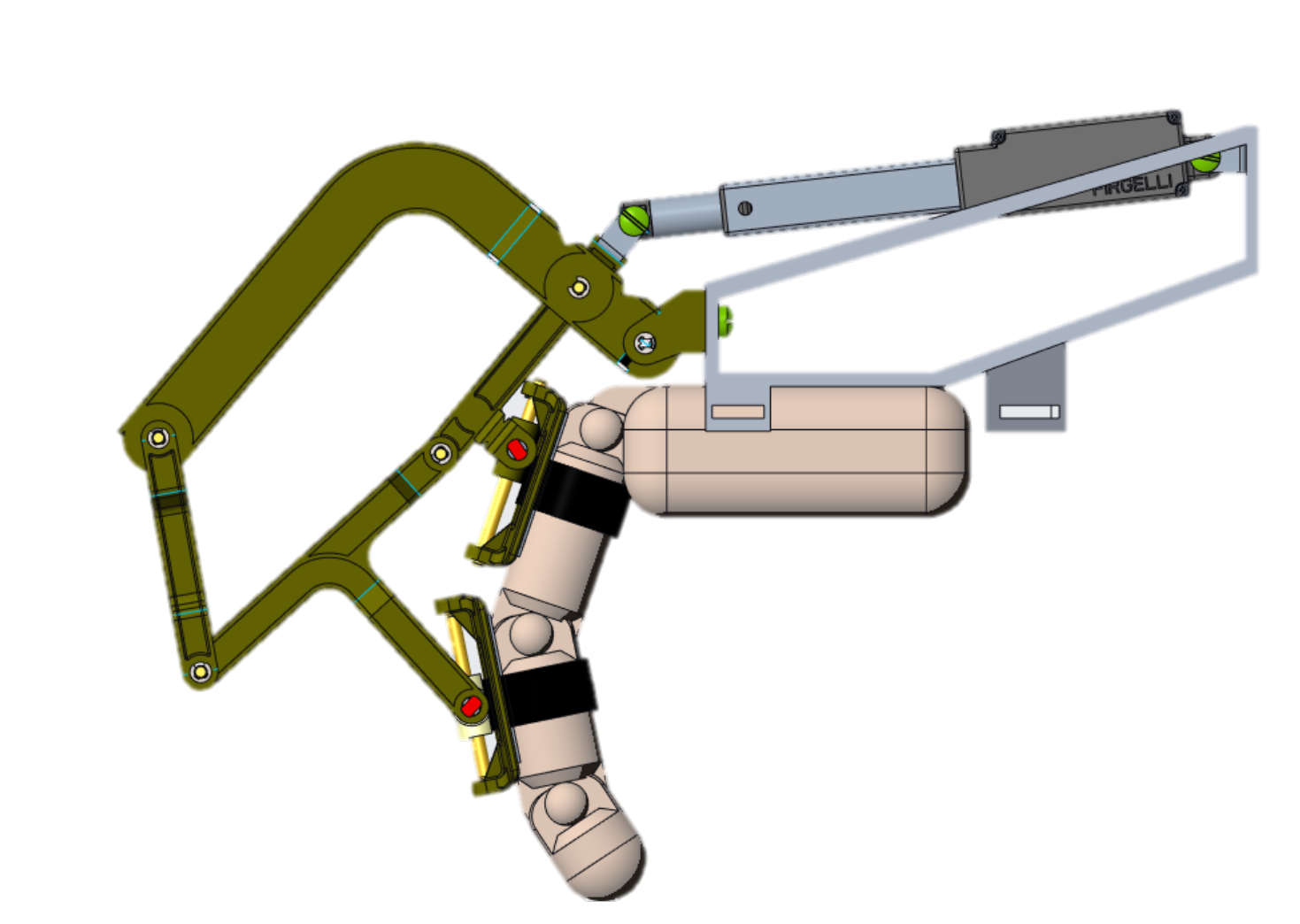}}
    \caption{CAD models of the designs retrieved by the optimization process, chosen among designs having $L_x \leq 50$ to comply with U-HEx's linear actuator (Firgelli L12-P). Specifically, (a) the design from the SOOP along with three designs from the MOOP: with reference to Fig.~\ref{fig:pareto}, (b) Design 5 with minimum actuator displacement ($obj_3$), (c) Design 6 with maximum force transmission ($obj_1$), (d) and Design 7 as a balanced trade-off among all objectives.}
    \label{fig:cad_models}
\end{figure*}

\section{Comparison of Optimized Robotic Designs} \label{sec:comparison}

The optimization process provided us with two designs from SOOP (with $L_x \leq 50$ and with $L_x > 50$, respectively) and 3665 designs from MOOP -- when considering nine link lengths as decision variables. We selected five of these 3665 trade-off designs (i.e., equally optimal) based on our preferences. The objective values and the numerical link lengths of these seven designs are reported in Table~\ref{tab:best_designs_values} and graphically crossed on the objective space in Fig.~\ref{fig:pareto} -- including 1 and 2 from SOOP for comparison. Additionally, we drew the CAD models for designs having $L_x \leq 50$ mm to comply with U-Hex's linear actuator (Frigelli L12-P), depicted in Fig.~\ref{fig:cad_models}. Design 1 and 2 were retrieved by BBBC, and they are the best solutions considering only force transmission as objective, whereas the MOOP designs were chosen based on the following criteria: Design 3 features the maximum force transmission (by SPEA2), Design 4 features the minimum in torque variance (by SPEA2), Design 5 features the minimum actuator displacement (by NS-BBBC), Design 6 features the maximum force transmission with an actuator displacement $L_x \leq 50$ (by SPEA2), and 
Design 7 features a good trade-off that balances all objectives (i.e., we selected a design with a low torque variance and a short actuator displacement with a ``not too low'' force transmission).

Fig.~\ref{fig:pareto} clearly shows that the solutions from SOOP (Design 1 and 2) are outside the Pareto set retrieved by the MOEAs; additionally, they are both non-dominated when compared to the rest of the set. Therefore, the true Pareto Front is actually larger than what was explored by the MOEAs. However, this does not mean that the MOEAs perform badly, but that the algorithms were attracted to specific regions of the objective space. This is clearly visible when observing the two-dimensional projections: Fig.~\ref{fig:pareto_13} shows that most solutions are concentrated in what would be the Pareto Front if $obj_1$ was not considered (i.e., solutions that minimize both $obj_2$ and $obj_3$). In fact, Design 1 and 2 lie in the opposite regions, but are still non-dominated because of the additional third objective $obj_1$. The opposite trend is visible also in the projection of Fig.~\ref{fig:pareto_12}: most of the solutions are not concentrated on what would be the Pareto Front if $obj_3$ was not considered (i.e., solutions that minimize both $obj_1$ and $obj_2$, whereas $obj_1$ was supposed to be maximized), indicating that overall $obj_3$ has a bigger impact than $obj_1$ on the objective space exploration.
From this figure, we can clearly observe that Design 1 lies on what would be the Pareto Front if $obj_3$ was not considered, whereas Design 2 lies on what would be the Pareto Front if also the constraint on the actuator $con_7$ is considered. From these observations, we can conclude that when considering force transmission only, Design 1 and 2 are overall the best solutions, indicating that BBBC for SOOP is a valid EA for robotic design optimization.


When comparing Design 3 to Design 1 (both $L_x > 50$), Design 3 renders lower forces, but it features smaller torque variance and a smaller actuator (therefore, the solutions do not dominate each other); however, the gain in the last objectives is so small that it might not justify the gain in the first objective. Design 4 (also $L_x > 50$) has notably better torque variance than Design 1 and 3, as expected, but a lower force transmission. As shown by Fig.~\ref{fig:pareto_23}, achieving a more balanced torque spread on the finger joints requires longer actuators -- and, therefore, lower force transmission.

Among the designs with $L_x \leq 50$, depicted in Fig.~\ref{fig:cad_models}, Design 5 is the one with the lowest actuator displacement. However, since we observed an inversely proportional relationship between force and actuator (see Fig.~\ref{fig:pareto_13}, Design 5 features the lowest force transmission among all possible solutions; additionally, its torque variance is high, resulting in unbalanced forces on the finger joints -- which are still in a safe range for the user. For Design 6, we picked a trade-off solution considering force transmission and actuator displacement with no particular attention to the torque variance: specifically, we chose the solution with the maximum force transmission having $L_x \leq 50$. This design resulted in very similar values of force transmission and actuator displacements to Design 2 from SOOP; however, it has a lower torque variance -- which is, surprisingly, the objective we did not consider while picking Design 4. This shows how MOOP provided a more optimized solution than SOOP, provided that designers are not interested in force transmission only. Finally, we picked Design 7 as a generic trade-off solution considering all the objectives, which produces lower forces than other designs but provides better torque balance and smaller actuators.


All solutions in Fig.~\ref{fig:pareto} are equally suitable depending on the designer's preference and, more specifically, based on the type of applications and use cases. U-HEx was previously proposed to be used not only for physical rehabilitation exercises but also for haptic or assistive uses. For haptic use, achieving smaller and lighter designs (therefore smaller actuator displacements) should be a higher priority than torque variance or force transmission (i.e., Design 5). For physical rehabilitation, achieving maximum force transmission should be a higher priority than torque variance or actuator displacements to overcome high levels of spasticity of their fingers among the feasible solutions (i.e., Design 6). Finally, for assistive use, all objectives need to be optimized simultaneously: high-force transmission is needed to help users grasp objects at any weight in a stable manner, and low torque variance is needed to improve their safety while grasping objects of any shape or size, and low actuator displacement is needed to improve the portability (i.e., Design 7). Compared to Design 2 in SOOP, which offers a very similar solution to Design 6, the solutions in MOOP offer designers the luxury to select \textit{an optimal} solution in parallel to their design needs. 

The physical robot shown in Fig.~\ref{fig:uhex_new} (first pages) is Design 6.

\section{Conclusions}
\label{sec:conclusion}

In this work, we presented a complete study on the optimization of U-HEx, an underactuated hand exoskeleton with a complex kinematic structure. We implemented two optimization design processes to find the correct link lengths that maximize the force transmission while: (1) ensuring the device's safe and feasible operation through mechanical constraints and (2) minimizing torque variance and the desired actuator displacement. 
While combining different multi-optimization survival strategies, we implemented two new MOEAs, namely NS-BBBC and SP-BBBC, which performed competitively with respect to two of the most famous MOEAs in the state-of-the-art, NSGA-II~\cite{deb2002fast} and SPEA2~\cite{zitzler2001spea2}. 
\textit{In terms of robotic designs}, our results indicate the potential benefits of different optimization problem settings depending on the requirements. 
It is interesting to observe that implementing a SOOP and MOOP with the same additional constraint ($con_7$) resulted in somewhat similar solutions, therefore the designer should make their selection among MOOP and SOOP considering the trade-off between flexibility of chosen design and computational burden. \textit{In terms of algorithms}, our 
results indicated that GA is more suitable for designers needing diverse conditions, while BBBC provides more consistent results for those less familiar with optimization methods.

In the future, this study can be extended by using another survival strategy for MOOPs: ``rank partitioning'', which was developed to optimize the design of soft growing robots~\cite{stroppa2024design}. This method retrieves a single optimal solution rather than a Pareto set once the designer has established priorities among objectives -- with no need for numerical values for each priority. Since the order of the objectives can produce a different design for specific applications, this method can quickly determine the targeted settings. Besides exploring different optimization methods, we will also examine the effects of the proposed improvements during real human-robot interaction by testing the physical exoskeletons. This investigation will include a thorough analysis of user experience and interaction forces to understand the practical impact and effectiveness of each design in enhancing human-robot collaboration.

\section*{Acknowledgments}
This work is funded by TUB\.ITAK project number 123M690 and partially funded by TUB\.ITAK project number 121C145 and 121C147.

The authors would like to thank Mazhar Eid Zyada for drawing the CAD models, and Ayse Yalcin and Ayodele Oyejide for printing and assembling the physical device.

\section*{Supplementary Information}

The MATLAB script is fully available at \url{https://www.mediafire.com/file/pwvzb1u6nw4k9e5/soft-robots-studio-3d.zip/file}.

\bibliographystyle{elsarticle-num} 
\bibliography{references}

\begin{thebibliography}{10}
\expandafter\ifx\csname url\endcsname\relax
  \def\url#1{\texttt{#1}}\fi
\expandafter\ifx\csname urlprefix\endcsname\relax\def\urlprefix{URL }\fi
\expandafter\ifx\csname href\endcsname\relax
  \def\href#1#2{#2} \def\path#1{#1}\fi

\bibitem{poliero2020applicability}
T.~Poliero, M.~Lazzaroni, S.~Toxiri, C.~Di~Natali, D.~G. Caldwell, J.~Ortiz, Applicability of an active back-support exoskeleton to carrying activities, Frontiers in Robotics and AI 7 (2020) 579963.

\bibitem{butzer2021fully}
T.~B{\"u}tzer, O.~Lambercy, J.~Arata, R.~Gassert, Fully wearable actuated soft exoskeleton for grasping assistance in everyday activities, Soft robotics 8~(2) (2021) 128--143.

\bibitem{stroppa2017robot}
F.~Stroppa, C.~Loconsole, S.~Marcheschi, A.~Frisoli, A robot-assisted neuro-rehabilitation system for post-stroke patients’ motor skill evaluation with alex exoskeleton, in: Proceedings of the International Conference on NeuroRehabilitation (ICNR), 2017, pp. 501--505.

\bibitem{huang2015design}
J.~Huang, X.~Tu, J.~He, Design and evaluation of the rupert wearable upper extremity exoskeleton robot for clinical and in-home therapies, IEEE Transactions on Systems, Man, and Cybernetics: Systems 46~(7) (2015) 926--935.

\bibitem{pons2010rehabilitation}
J.~L. Pons, Rehabilitation exoskeletal robotics, IEEE Engineering in Medicine and Biology Magazine 29~(3) (2010) 57--63.

\bibitem{sarac2019design}
M.~Sarac, M.~Solazzi, A.~Frisoli, Design requirements of generic hand exoskeletons and survey of hand exoskeletons for rehabilitation, assistive, or haptic use, IEEE Transactions on Haptics (ToH) 12~(4) (2019) 400--413.

\bibitem{van2015considerations}
J.~Van~der Vorm, L.~O'Sullivan, R.~Nugent, M.~de~Looze, Considerations for developing safety standards for industrial exoskeletons, Robo-Mate (2015) 1--13.

\bibitem{fisahn2016effectiveness}
C.~Fisahn, M.~Aach, O.~Jansen, M.~Moisi, A.~Mayadev, K.~T. Pagarigan, J.~R. Dettori, T.~A. Schildhauer, The effectiveness and safety of exoskeletons as assistive and rehabilitation devices in the treatment of neurologic gait disorders in patients with spinal cord injury: A systematic review, Global spine journal 6~(8) (2016) 822--841.

\bibitem{sioshansi2017optimization}
R.~Sioshansi, A.~J. Conejo, Optimization in Engineering, Vol. 120, Cham: Springer International Publishing, 2017.

\bibitem{statnikov2012multicriteria}
R.~B. Statnikov, J.~B. Matusov, Multicriteria Optimization and Engineering, Springer Science and Business Media, 2012.

\bibitem{andersson2000survey}
J.~Andersson, A survey of multiobjective optimization in engineering design, Department of Mechanical Engineering, Linktjping University. Sweden (2000).

\bibitem{bonnans2006numerical}
J.-F. Bonnans, J.~C. Gilbert, C.~Lemar{\'e}chal, C.~A. Sagastiz{\'a}bal, Numerical Optimization: Theoretical and Practical Aspects, Springer Science and Business Media, 2006.

\bibitem{stroppa2023optimizing}
F.~Stroppa, A.~Soylemez, H.~T. Yuksel, B.~Akbas, M.~Sarac, Optimizing exoskeleton design with evolutionary computation: An intensive survey, Robotics 12~(4) (2023) 106.

\bibitem{norde2000characterizing}
H.~Norde, F.~Patrone, S.~Tijs, Characterizing properties of approximate solutions for optimization problems, Mathematical Social Sciences 40~(3) (2000) 297--311.

\bibitem{dizangian2015reliability}
B.~Dizangian, M.~Ghasemi, Reliability-based design optimization of complex functions using self-adaptive particle swarm optimization method, International Journal of Optimization in Civil Engineering 5~(2) (2015) 151--165.

\bibitem{nomaguchi2016robust}
Y.~Nomaguchi, K.~Kawakami, K.~Fujita, Y.~Kishita, K.~Hara, M.~Uwasu, Robust design of system of systems using uncertainty assessment based on lattice point approach: Case study of distributed generation system design in a japanese dormitory town, International Journal of Automation Technology 10~(5) (2016) 678--689.

\bibitem{dumitrescu2000evolutionary}
D.~Dumitrescu, B.~Lazzerini, L.~C. Jain, A.~Dumitrescu, Evolutionary Computation, CRC press, 2000.

\bibitem{wang2017autonomous}
W.~Wang, D.~Gu, G.~Xie, Autonomous optimization of swimming gait in a fish robot with multiple onboard sensors, IEEE Transactions on Systems, Man, and Cybernetics: Systems 49~(5) (2017) 891--903.

\bibitem{datta2015analysis}
R.~Datta, S.~Pradhan, B.~Bhattacharya, Analysis and design optimization of a robotic gripper using multiobjective genetic algorithm, IEEE Transactions on Systems, Man, and Cybernetics: Systems 46~(1) (2015) 16--26.

\bibitem{stroppa2024optimizing}
F.~Stroppa, F.~J. Majeed, J.~Batiya, E.~Baran, M.~Sarac, Optimizing soft robot design and tracking with and without evolutionary computation: an intensive survey, Cambridge University Press Robotica (2024) 1--37.

\bibitem{sarac2017design}
M.~Sarac, M.~Solazzi, E.~Sotgiu, M.~Bergamasco, A.~Frisoli, Design and kinematic optimization of a novel underactuated robotic hand exoskeleton, Meccanica 52 (2017) 749--761.

\bibitem{akbas2024impact}
B.~Akbas, H.~T. Yuksel, A.~Soylemez, M.~E. Zyada, M.~Sarac, F.~Stroppa, The impact of evolutionary computation on robotic design: A case study with an underactuated hand exoskeleton, in: IEEE International Conference on Robotics and Automation (ICRA), 2024.

\bibitem{goldberg1987genetic}
D.~E. Goldberg, J.~Richardson, et~al., Genetic algorithms with sharing for multimodal function optimization, in: Genetic Algorithms and Their Applications: Proceedings of the Second International Conference on Genetic Algorithms, Vol. 4149, 1987.

\bibitem{goldberg1990real}
D.~E. Goldberg, et~al., Real-coded genetic algorithms, virtual alphabets and blocking, Citeseer, 1990.

\bibitem{erol2006new}
O.~K. Erol, I.~Eksin, A new optimization method: Big {B}ang--{B}ig {C}runch, Advances in Engineering Software 37~(2) (2006) 106--111.

\bibitem{Denizon2024}
D.~Denizon, S.~Dogru, L.~Marques, Improving grasping performance of underactuated two finger robotic hands using variable stiffness, in: IEEE Iberian Robotics Conference (ROBOT), 2024, pp. 1--6.
\newblock \href {https://doi.org/10.1109/ROBOT61475.2024.10797355} {\path{doi:10.1109/ROBOT61475.2024.10797355}}.

\bibitem{Yan2022}
Y.~Yan, S.~Guo, C.~Yang, C.~Lyu, L.~Zhang, The pg2 gripper: an underactuated two-fingered gripper for planar manipulation, in: IEEE International Conference on Mechatronics and Automation (ICMA), 2022, pp. 680--685.
\newblock \href {https://doi.org/10.1109/ICMA54519.2022.9856375} {\path{doi:10.1109/ICMA54519.2022.9856375}}.

\bibitem{Li2023}
H.~Li, L.~Han, X.~Xia, Z.~Liu, Y.~Zhang, Y.~Wang, L.~Cheng, Design of an assistive hand exoskeleton based on the principles of the human hand grasp, in: IEEE International Conference on Mechatronics and Machine Vision in Practice (M2VIP), 2023, pp. 1--6.
\newblock \href {https://doi.org/10.1109/M2VIP58386.2023.10413419} {\path{doi:10.1109/M2VIP58386.2023.10413419}}.

\bibitem{Gu2024}
S.~Gu, Z.~Ye, L.~Zhang, R.~Peng, J.~Wang, H.~Li, Research on a novel hand exoskeleton rehabilitation training system, in: IEEE International Conference on Mechatronics and Automation (ICMA), 2024, pp. 496--501.
\newblock \href {https://doi.org/10.1109/ICMA61710.2024.10632949} {\path{doi:10.1109/ICMA61710.2024.10632949}}.

\bibitem{goldberg1989genetic}
D.~E. Goldberg, Genetic Algorithms in Search, Optimization, and Machine Learning, Addison-Wesley, 1989.

\bibitem{goldberg1991comparative}
D.~E. Goldberg, K.~Deb, A comparative analysis of selection schemes used in genetic algorithms, in: Foundations of genetic algorithms, Vol.~1, Elsevier, 1991, pp. 69--93.

\bibitem{eshelman1993real}
L.~J. Eshelman, J.~D. Schaffer, Real-coded genetic algorithms and interval-schemata, in: Foundations of genetic algorithms, Vol.~2, Elsevier, 1993, pp. 187--202.

\bibitem{deb1996combined}
K.~Deb, M.~Goyal, et~al., A combined genetic adaptive search (geneas) for engineering design, Computer Science and informatics 26 (1996) 30--45.

\bibitem{beyer2002evolution}
H.-G. Beyer, H.-P. Schwefel, Evolution strategies -- a comprehensive introduction, Natural computing 1 (2002) 3--52.

\bibitem{gencc2010big}
H.~M. Gen{\c{c}}, I.~Eksin, O.~K. Erol, Big bang-big crunch optimization algorithm hybridized with local directional moves and application to target motion analysis problem, in: IEEE International Conference on Systems, Man and Cybernetics, 2010, pp. 881--887.

\bibitem{coello2002theoretical}
C.~A.~C. Coello, Theoretical and numerical constraint-handling techniques used with evolutionary algorithms: A survey of the state of the art, Computer Methods in Applied Mechanics and Engineering 191~(11-12) (2002) 1245--1287.

\bibitem{deb2001multi}
K.~Deb, Multi-objective Optimization using Evolutionary Algorithms, John Wiley and Sons, Ltd, 2001.

\bibitem{deb2002fast}
K.~Deb, A.~Pratap, S.~Agarwal, T.~Meyarivan, A fast and elitist multiobjective genetic algorithm: {NSGA-II}, IEEE Transactions on Evolutionary Computation 6~(2) (2002) 182--197.

\bibitem{zitzler2001spea2}
E.~Zitzler, M.~Laumanns, L.~Thiele, {SPEA2}: Improving the strength pareto evolutionary algorithm, TIK-report 103 (2001).

\bibitem{shang2020survey}
K.~Shang, H.~Ishibuchi, L.~He, L.~M. Pang, A survey on the hypervolume indicator in evolutionary multiobjective optimization, IEEE Transactions on Evolutionary Computation 25~(1) (2020) 1--20.

\bibitem{fonseca2006improved}
C.~M. Fonseca, L.~Paquete, M.~L{\'o}pez-Ib{\'a}nez, An improved dimension-sweep algorithm for the hypervolume indicator, in: IEEE International Conference on Evolutionary Computation, 2006, pp. 1157--1163.

\bibitem{while2006faster}
L.~While, P.~Hingston, L.~Barone, S.~Huband, A faster algorithm for calculating hypervolume, IEEE Transactions on Evolutionary Computation 10~(1) (2006) 29--38.

\bibitem{king2010comparison}
R.~A. King, K.~Deb, H.~Rughooputh, Comparison of {NSGA-II} and {SPEA2} on the multiobjective environmental/economic dispatch problem, University of Mauritius Research Journal 16~(1) (2010) 485--511.

\bibitem{stroppa2024design}
F.~Stroppa, Design optimizer for planar soft-growing robot manipulators, Elsevier Engineering Applications of Artificial Intelligence 130 (2024) 107693.

\end{thebibliography}

\end{document}